\documentclass[runningheads]{llncs}
\usepackage{graphicx}
\usepackage{comment}
\usepackage{amsmath,amssymb} % define this before the line numbering.
\usepackage{color}
\usepackage[colorlinks]{hyperref}

\makeatletter
\DeclareRobustCommand\onedot{\futurelet\@let@token\@onedot}
\def\@onedot{\ifx\@let@token.\else.\null\fi\xspace}

\def\wrt{w.r.t\onedot} 
\def\etal{\emph{et al}\onedot}
\makeatother

\usepackage{microtype}
\usepackage{epsfig}
\usepackage{graphicx}
\usepackage{amsmath}
\usepackage{amssymb}
\usepackage{color}
\usepackage{caption}
\usepackage{booktabs}

\usepackage{wrapfig}
\usepackage{multirow}
\usepackage{gensymb}
\usepackage{xspace}

\definecolor{turquoise}{cmyk}{0.65,0,0.1,0.3}
\definecolor{purple}{rgb}{0.65,0,0.65}
\definecolor{dark_green}{rgb}{0, 0.5, 0}
\definecolor{orange}{rgb}{0.8, 0.6, 0.2}
\definecolor{red}{rgb}{0.8, 0.2, 0.2}
\definecolor{blueish}{rgb}{0.0, 0.7, 1}
\definecolor{light_gray}{rgb}{0.7, 0.7, .7}
\definecolor{pink}{rgb}{1, 0, 1}
\definecolor{dark_red}{rgb}{0.5, 0, 0}

\usepackage{pifont}
\usepackage{xcolor}
\newcommand{\cmark}{\textcolor{green!80!black}{\ding{51}}}
\newcommand{\xmark}{\textcolor{red}{\ding{55}}}

\newcommand{\tS}{\mbox{$\tilde S$}}
\newcommand{\tI}{\mbox{$\tilde I$}}

\begin{document}
% \linenumbers
\pagestyle{headings}
\mainmatter
\def\ECCVSubNumber{2305}  % Insert your submission number here

%%%%%%%%% TITLE
\title{Shape and Viewpoint without Keypoints}

%\maketitle
%\thispagestyle{empty}

\begin{comment}
\titlerunning{ECCV-20 submission ID \ECCVSubNumber} 
\authorrunning{ECCV-20 submission ID \ECCVSubNumber} 
\author{Anonymous ECCV submission}
\institute{Paper ID \ECCVSubNumber}
\end{comment}
%******************

% \begin{comment}
\titlerunning{U-CMR}
% \author{Shubham Goel\inst{1}\orcidID{0000-1111-2222-3333} \and
% Angjoo Kanazawa\inst{1}\orcidID{1111-2222-3333-4444} \and
% Jitendra Malik\inst{1}\orcidID{2222--3333-4444-5555}}
\author{Shubham Goel \and
Angjoo Kanazawa \and
Jitendra Malik}
\authorrunning{S. Goel et al.}
\institute{UC Berkeley}
% \end{comment}
%******************
\maketitle
%%%%%%%%% ABSTRACT
\begin{abstract}
We present a learning framework that learns to recover the 3D shape, pose and texture from a single image, trained on an image collection without any 
ground truth 3D shape, multi-view, camera viewpoints or keypoint supervision. 
We approach this highly under-constrained problem in a ``analysis by synthesis" framework where the goal is to predict the likely shape, texture and camera viewpoint that could produce the image with various learned category-specific priors. Our particular contribution in this paper is a representation of the distribution over cameras, which we call ``camera-multiplex".
Instead of picking a point estimate, we maintain a set of camera hypotheses that are optimized during training to best explain the image given the current shape and texture.  
We call our approach Unsupervised Category-Specific Mesh Reconstruction (U-CMR), and present qualitative and quantitative results on CUB, Pascal 3D and new web-scraped datasets. We obtain state-of-the-art camera prediction results and show that we can learn to predict diverse shapes and textures across objects using an image collection without any keypoint annotations or 3D ground truth. Project page: \footnotesize{\url{https://shubham-goel.github.io/ucmr}}

\end{abstract}

\begin{center}
    \newcommand{\teaserwidth}{\textwidth}
    % \vspace{-3mm}
    \centerline{
    \includegraphics[width=\teaserwidth]{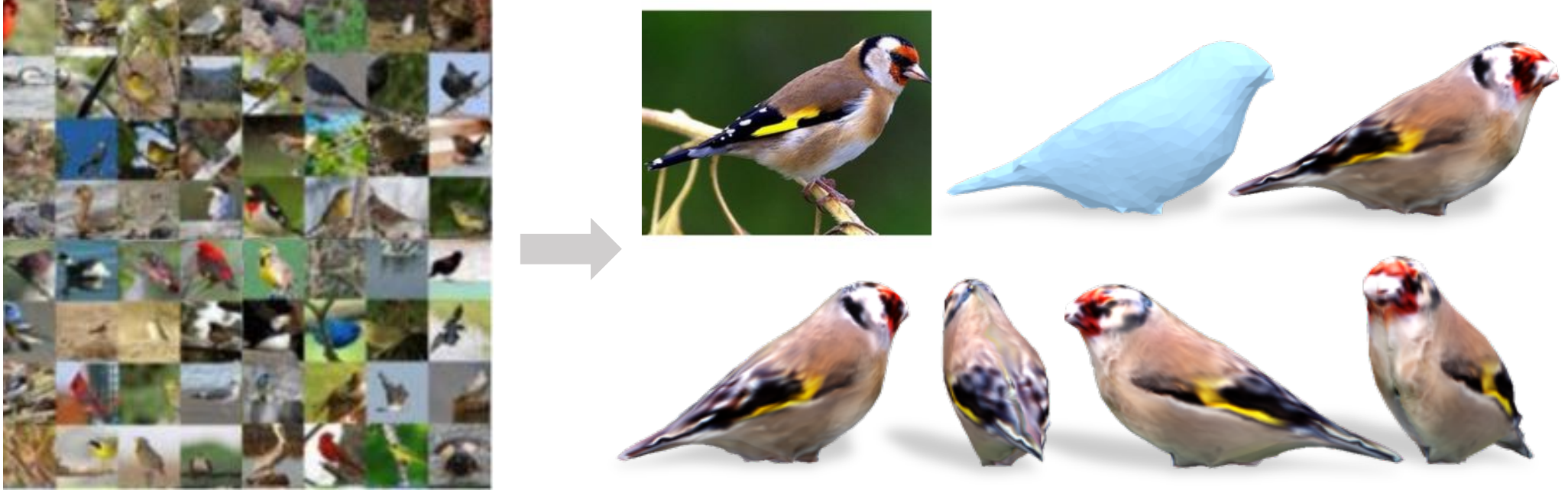}
     }
    \captionof{figure}{\small Given an image collection of an object category, like birds, we propose a computational framework that given a single image of an object, predicts its 3D shape, viewpoint and texture, without using any 3D shape, viewpoints or keypoint supervision during training. On the right we show the input image and the results obtained by our method, shown from multiple views.}
    % \vspace{-2mm}
	\label{fig:teaser}

\end{center}%

%%%%%%%%% BODY TEXT
\section{Introduction}

There has been much progress in recent years in training deep networks to infer 3D shape from 2D images. These approaches fall into two major families based on the supervisory signal used (a) 3D models, as available in collection of CAD models such as ShapeNet or (b) multiple views of the same object which permit  deep learning counterparts of classical techniques such as shape carving or structure-from-motion. But do we need all this supervision?

It is easy to find on the internet large collections of images of objects belonging to particular categories, such as birds or cars or chairs. Let us focus on birds, for which datasets like CUB \cite{Wah}, shown in Figure~\ref{fig:teaser} (left) exist. Note that this is a ``Multiple Instance Single View" setting. For each bird instance we have only a single view, and every bird is a slightly different shape, even though the multiple instances share a family resemblance. Compared to classical SFM, where we have ``Single Instance Multiple Views", our goal is to ``3Dfy" these birds.  From a single image, create a 3D model and its texture map, which can then be rendered from different camera viewpoints as shown in the rows of Figure~\ref{fig:teaser} (right). %This particular formulation was presented in the recent work of Kanazawa \etal \cite{cmrKanazawa18}, which is an inspiration for our work
This particular formulation was presented in the ``Category-Specific Mesh Reconstruction"  work of Kanazawa \etal \cite{cmrKanazawa18}, and their algorithm (CMR) is an inspiration for our work. %\sg{ an inspiration for our work / a predecessor to our work / is what we build upon}. 
Even earlier the work of Cashman and Fitzgibbon \cite{Cashman} working on analyzing images of dolphins showed how an ``analysis by synthesis" paradigm with a deformable template model of a shape category could enable one to infer 3D shapes in an optimization framework.

It is under-appreciated that these approaches, while pioneering, do exploit some supervisory information. This includes (1) knowledge of a mean shape for the category (2) silhouettes for each instance (3) marked  keypoints on the various instances (e.g. beak tip for each bird). Of these, the need for labeled keypoints is the most troublesome. Only one mean shape is needed for the entire category and sometimes a very generic initialization such as a sphere is good enough. 
Silhouettes could presumably be marked (perhaps in a category-agnostic way) by an instance segmentation network like Mask R-CNN. But keypoints are tedious to mark on every instance. This effort can be justified for a single important category like humans, but we cannot afford to do for the thousands of categories which we might want to ``3Dfy". Yes, we could have the keypoints be marked by a network but to train that would require keypoint labels! There have been recent efforts in unsupervised keypoint-detection \cite{jakab2018unsupervised}, however, so far, these methods learn viewpoint-dependent keypoints that often get mixed-up in the presence of 180-degree rotations.

In this paper we present an approach, U-CMR (for Unsupervised CMR) which enables us to train a function which can factor an image into a 3D shape, texture and camera, in a roughly similar setting to CMR \cite{cmrKanazawa18}, except that we replace the need for keypoint annotation with a single 3D template shape for the entire category. It turns out that keypoint annotations are needed for recovering cameras (using SFM like techniques) and if we don't have keypoint annotations, we have to solve for the camera simultaneously with shape and texture. Extending the ``analysis by synthesis" paradigm to also recover cameras is unfortunately rather hard.  Intuitively speaking, this is because of the discontinuous and multi-modal nature of the space of possible camera viewpoints. While shape and texture have a smooth, well behaved optimization surface that is amenable to gradient descent optimization, this does not hold for the space of possible cameras. The two most likely camera explanations might lie on the opposite sides of the viewing sphere, where it is not possible to approach the optimal camera in an iterative manner as done when optimizing the energy landscape of a deep network. This typically causes the camera prediction to be stuck in bad local minima. Our solution to this problem is to maintain a set of possible camera hypotheses for each training instance that we call a \textbf{camera-multiplex}. This is reminiscent of particle filtering approaches with the idea being to maintain a distribution rather than prematurely pick a point estimate. We can iteratively refine the shape, texture as well as the camera-multiplex. More details are presented in Section \ref{sec:approach}.

We evaluate our approach on 3D shape, pose, and texture reconstruction on 4 categories: CUB-200-2011 birds \cite{Wah}, PASCAL-3D \cite{xiang2014beyond} cars, motorcycles and a new dataset of shoes we scraped from the Internet. 
We show that naively predicting shape, camera, and texture directly results in a degenerate solution where the shape is flat and the camera collapses to a single mode. We quantitatively evaluate the final camera prediction where the proposed camera-multiplex approach obtains the state of the art results under this weakly-supervised setting of no keypoint annotations.  We show that despite the lack of viewpoints and keypoints, we can learn a reasonable 3D shape space, approaching that of the shapes obtained by a previous method that uses keypoint supervision \cite{cmrKanazawa18}.

\section{Related Work}
% Deep learning based methods that predict 3D shape
% When deep learning approaches are developed for 3D shape
Recent deep learning based 3D reconstruction methods can be categorized by the required supervisory signal and the output of the system. This is illustrated in Table 1. 
%P1: 3D supervision
Earlier methods formulate the problem assuming full 3D shape supervision for an image \cite{choy20163d,girdhar16b,psgn,factored3dTulsiani17,meshrcnn}, which is enabled by synthetic datasets such as ShapeNet \cite{wu20153d} and SunCG \cite{song2017semantic}. Some approaches generate synthetic datasets using data gathered from the real world to train their models  \cite{chen2016synthesizing,varol17_surreal,Zuffi19Safari}. However, requiring 3D supervision severely restricts these approaches, since ground truth 3D shape is costly or not possible to acquire, especially at a large scale.  %While 3D supervision is available in synthetic datasets, their availability is severely limited from real images.
As such, follow up methods explore more natural forms of supervision, where multiple-views of the same object are available. Some of these approaches assume known viewpoints \cite{yan2016perspective,lsmKarHM2017,drcTulsiani17} akin to the setting of traditional MVS or visual hull. Other approaches explore the problem with unknown viewpoint setting \cite{mvcTulsiani18,insafutdinov2018unsupervised,gadelha20173d}. These approaches assume that multi-view silhouettes, images, and/or depth images are available, and train their models such that the predicted 3D shapes reconstruct the images after projection or differentiable rendering. A variety of differentiable rendering mechanisms have been explored \cite{loper2014opendr,NMR,liu2019soft}. 

While multi-view images may be obtained in the real world, the vast amount of available visual data corresponds to the setting of unconstrained collection of single-view images, where no simultaneous multiple views of the same instance are available. This is also the natural setting for non-rigid objects where the shape may change over time. The traditional non-rigid structure from motion  \cite{torresani2008nonrigid} also falls under this category, where the input is a tracked set of corresponding points \cite{torresani2008nonrigid,dai2014simple} or 2D keypoints \cite{Vincente,novotny2019c3dpo}. Earlier approaches fit a deformable 3D model \cite{BlanzVetter,CSDM,Cashman,kanazawa2016learning} to 2D keypoints and silhouettes. Kanazawa \etal \cite{cmrKanazawa18} propose CMR, a learning based framework where 3D shape, texture, and camera pose are predicted from a single image, trained under this setting of single-view image collections with known mask and keypoint annotations. While this is a step in the right direction, the requirement of keypoint annotation is still restrictive. More recently, Kulkarni \etal \cite{kulkarni2019csm} bypass this requirement of keypoints to learn a dense canonical surface mapping of objects from a set of image collections with mask supervision and a template 3D shape. They focus on predicting the surface correspondences on images and learn to predict the camera viewpoints during the training, but do not learn to predict the 3D shape. While we tackle a different problem, we operate under the same required supervision. As such we quantitatively compare with CSM on the quality of the camera predicted, where our approach obtains considerably better camera predictions. %.over the recent approach of Kulkarni \etal~\cite{kulkarni2019csm}, which also learns to predict the camera (but not the shape) without any camera viewpoint, or keypoints supervision.
Note that there are several recent approaches that explore disentangling images into 2.5D surface properties, camera, and texture of the visible regions from a collection of monocular images, without any masks \cite{thewlis2017unsupervised,shu2018deforming,wu2019}. However these approaches are mainly demonstrated on faces. In this work we recover a full 3D representation and texture from a single image. 

\begin{center}
\begin{table*}[!t]
% \vspace{-0.5em}
\resizebox{\textwidth}{!}{
%\begin{tabular}{c|c|c|c|c|c|c|c|c}
\begin{tabular}{c|ccccc|cccc}
\toprule%\hline
Approach & \multicolumn{5}{c}{Required Supervision per Image} & \multicolumn{4}{|c}{Output}					\\ 
{} & 3D Shape & Multi-view & Cam & Keypoints & Mask & 3D Shape & 2.5D & Cam & Texture \\
\midrule
MeshRCNN$^*$ \cite{meshrcnn} &	\xmark	& & &	    &	    &   \cmark  &  &       &	    \\
DeepSDF \cite{park2019deepsdf} &	\xmark	& & &	    &	    &   \cmark  &  &       &	    \\
Smalst \cite{Zuffi19Safari}	      &	\xmark	& &	\xmark	&	\xmark &	\xmark	&	\cmark	&	& \cmark	&	\cmark	\\
PTN \cite{yan2016perspective} & & \xmark & \xmark & & & &  & \cmark &  \\
MVC \cite{mvcTulsiani18} & & \xmark &  & & & \cmark &  & \cmark & \\
CMR	 \cite{cmrKanazawa18} &				&		&	\xmark	&	\xmark	&	\xmark	&	\cmark	&	& \cmark	&	\cmark	\\
CSM \cite{kulkarni2019csm}	&			&		&		&		&	\xmark	&		&	& \cmark	&		\\
Wu \etal \cite{wu2019}	&			&		&		&		&	&		&	\cmark & \cmark	&	\cmark	\\
 \textbf{U-CMR}	&	&	&	&	&	\xmark	&	\cmark	& &	\cmark	&	\cmark	\\
% %\hline
\bottomrule
\end{tabular}
}
\caption{\small A comparison of different approaches highlighting the differences between the input (during training) and the output (during inference). Our approach (U-CMR) uses only silhouette supervision but predicts full 3D shape, camera and texture. 
{$^*$MeshRCNN predicts shape in camera-coordinates instead of a canonical frame.}
}
\label{table:inputoutput}
% \vspace{-2em}
\end{table*}
\end{center}

\section{Approach}
\label{sec:approach}
\begin{figure*}[!ht]
  \centering
  \includegraphics[width=\textwidth]{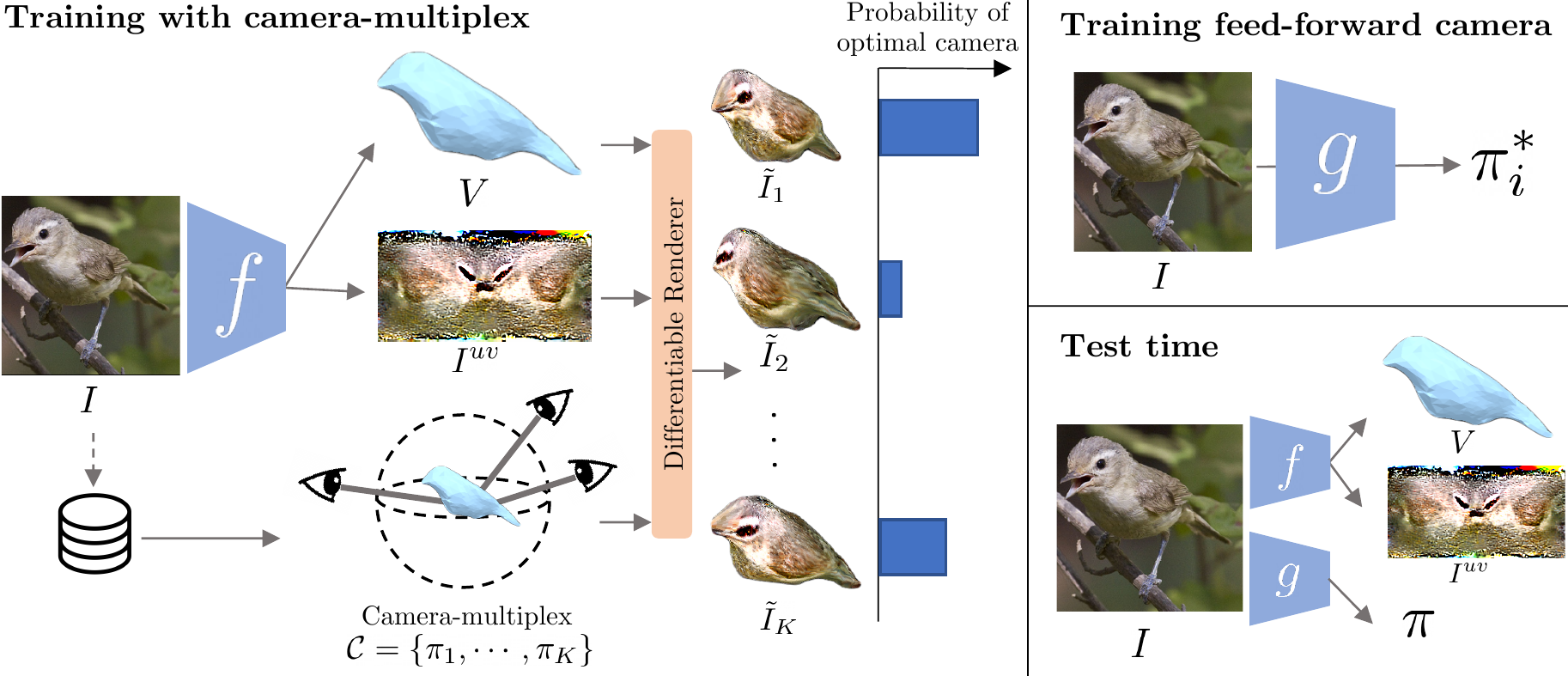}
  \caption{\small{\bf Overview of the proposed framework.} 
We first train a shape and texture predictor $f$, while simultaneously optimizing the camera-multiplex $\mathcal{C}$, a set of $K$ possible camera hypotheses maintained for every image. We render the predicted shape and texture from every camera in the multiplex, compute a per-camera reconstruction loss, treat it as negative log-likelihood of the camera, and update the $f$ against the expected loss. We also update each camera in the multiplex against the loss incurred by it.
After training $f$, we train a feed-forward model $g$ to predict the best camera $\pi^*$ in the multiplex from an image. As such, at test time our approach is able to predict all shape, texture, and camera from a single image.
  }
%   \vspace{-1em}
  \label{fig:overview}
\end{figure*}
% Our goal is  to learn to predict the 3D shape and texture of an object instance from a single image without using ground truth 3D shape or camera viewpoints or keypoint supervision. We solve this under-constrained problem by proposing a camera-multiplex approach, which allows learning of shape and texture predictor against multiple possible camera hypotheses at once. Our approach can be trained on an image collection of an object category with only the silhouette mask as supervision, along with a single 3D template shape for the entire category. We first go over the preliminary problem setup and then discuss the details of our approach that uses the multiplex camera %.characterize the issue when there are no viewpoints or keypoint supervision, and then d

\subsection{Preliminaries}
% Our approach follows the analysis-by-synthesis paradigm for disentangling the shape, texture, and camera from a single image \cite{roberts1963machine,BlanzVetter,cmrKanazawa18}. The idea is to recover these factors such that when they are rendered together the resulting image explains the image evidence, in forms of silhouette masks and the image pixels. Below we describe the parametrization of each of these factors. 

%\noindent\textbf{Shape Representation.}
\paragraph{Shape Representation.}
We represent 3D shape as a mesh $M \equiv (V, F)$ with vertices $V \in \mathbb{R}^{|V|\times 3}$ and faces $F$. The set of faces $F$ defines the connectivity of vertices in the mesh and we assume it remains fixed. We choose a mesh topology that is homeomorphic to a sphere. %As $F$ is fixed, we loosely use $V$ to refer to both the mesh and it's vertices in the rest of the paper. 
We model the vertex positions of a deformable object as $V = \Delta_V + \bar{V}$, the summation of an instance-specific deformation $\Delta_V$ that is predicted from an image to a learned instance-independent mean shape $\bar{V}$ \cite{cmrKanazawa18}.  We initialize the mean shape with the template 3D mesh.  
This parameterization allows the model to learn the space of possible deformations for each category.

%\noindent\textbf{Texture Representation.}
\paragraph{Texture Representation.}
%We use a UV-image representation similar to CMR \cite{cmrKanazawa18}. 
As the topology of our mesh is fixed, we can use a UV image $I^{uv}$ representation to model the texture. The values in a UV image get mapped onto the surface via a fixed UV mapping. The UV mapping is either a spherical projection akin to unrolling a globe into a flat map~\cite{hughes2014computer}, or when that is not good enough, is a distortion-minimizing unwrap of a template mesh 
along manually defined seams 
computed using blender \cite{blender}. 
%Previous work explored predicting the values of $I^{uv}$ as texture-flow \cite{cmrKanazawa18}. %However, we show that when there are no viewpoints or keypoints, directly predicting the texture values.
%In this work, we directly predict the texture values in a decoder framework, as the spatial structure in the UV image helps disambiguate incorrect cameras, we will expand on this point in Section \ref{sec:texture}. %show that this is more beneficial when no viewpoints or keypoints are available below. 

\paragraph{Camera Projection.}
%\noindent\textbf{
We assume a weak-perspective camera projection, parametrized by scale $\textbf{s} \in \mathbb{R}$, translation $\textbf{t} \in\mathbb{R}^2$ and rotation $\textbf{R}$ (captured as Euler angles azimuth, elevation, cyclo-rotation $[\textit{az,el,cr}] \in\mathbb{R}^3$).  We use $\pi(P)$ to denote the projection of a set of 3D points $P$ onto the image coordinates via the weak-perspective projection defined by $\pi \equiv (s, \textbf{t}, \textbf{R})$. We denote the image rendered by composing all three factors as $\tilde{I} = \mathcal{R}(V, I^{uv}, \pi)$ and silhouette rendered just from the shape and camera as $\tilde{S} = \mathcal{R}(V,\pi)$ where $\mathcal{R}(\cdot)$ is a differentiable renderer. We denote a set of camera hypotheses kept for each image, a camera-multiplex $\mathcal{C} = \{\pi_1, \cdots, \pi_K\}$. We describe its training details below.

\subsection{Our Method}

Figure \ref{fig:overview} shows an overview of our approach. During training, we learn a function $f(I)$ to predict the 3D shape and texture of the object underlying image $I$. We optimize over the camera-multiplex for each instance in the training dataset instead of making a deterministic prediction. %We call this set of possible camera pose hypotheses a \textit{camera-multiplex}.
For every shape and texture prediction, we compute the loss from every camera in the camera-multiplex, which induces a distribution on the camera poses inside a multiplex. We then use the expected loss over the camera-multiplex to update $f(I)$. % - which is used for training shape and texture.
%Each camera in the camera-multiplex is optimized proportionally to its likelihood. The shape and texture predictions are optimized to minimize the expected loss incurred over all of the camera-multiplex. We optimize this camera multiplex to minimize the loss corresponding to each camera.
When the training of $f(\cdot)$ converges, we identify the optimal camera for each training example in the camera-multiplex. We then train a function $g(I)$ that predicts the optimal camera from a single image, such that at test time we can infer all shape, texture, and camera from a single image. % we compile information from the optimized camera-multiplexes into a function $g_{\theta'}(I)$  \sg{(better phrase?)} to predict the most likely camera pose for a novel image $I$. 
We provide the details for the training process below.

% \subsection{Training Process}
For each training instance $I$, let $\mathcal{C} = \{\pi_1, \cdots, \pi_K\}$ denote its camera-multiplex with $K$ cameras and $S$ its silhouette. Note that while we omit the subscript on training instances for brevity, \emph{every} instance maintains its own $\mathcal{C}$ independently. For every predicted shape $V=\bar{V} + \Delta V$ and texture $I^{uv}$, we compute the silhouette and image reconstruction loss against each camera $\pi_k$: % be the rendered silhouette and image respectively, and :% The losses we use are silhouette loss:% render the silhouette $\tS_k$ and $\tI_k$ from each camera $\pi_k$ in the camera-multiplex and compute the silhouette loss:
\begin{equation}
\label{eq:mask-loss}
L_{\text{mask},k} = ||S - \tS_k||_2^2 + \texttt{dt}(S)*\tS_k,
\end{equation} 
\begin{equation}
\label{eq:pixel-loss}
L_{\text{pixel},k} = \texttt{dist}(\tI_k \odot S, I \odot S, ),
\end{equation} 
$L_{\text{mask},k}$ is the silhouette loss where $\tS_k = \mathcal{R}(V, \pi_k)$ is the silhouette rendered from camera $\pi_k$, and $\texttt{dt}(S)$ is the uni-directional distance transform of the ground truth silhouette. 
$L_{\text{pixel},k}$ is the image reconstruction loss computed over the foreground regions where $\mathcal{R}(V, I^{uv}, \pi_k)$ is the rendered image from camera $\pi_k$. For this, we use the perceptual distance metric of Zhang \etal \cite{zhang2018perceptual}. To exploit the bilateral symmetry and to ensure symmetric texture prediction, we also render the mesh under a bilaterally symmetric second camera, and compute $L_{\text{pixel},k}$ as the average pixel loss from the two cameras. 

In addition to these losses, we employ a graph-laplacian smoothness prior on our shape
$L_{\text{lap}} = ||V_i - \frac{1}{|N(i)|}\sum_{j\in N(i)}V_j ||^2$ that penalizes vertices $i$ that are far away from the centroid of their adjacent vertices $N(i)$. For cars, motorcycles and shoes, we empirically observe better results using $L_{\text{lap}} = ||LV||_2$ where $L$ is the discrete Laplace-Beltrami operator that minimizes mean curvature \cite{pinkall1993computing}. For this, we construct $L$ once using the template mesh at the start of training.
% For some experiments, we also tried $L_{\text{lap}} = ||LV||_2$ where $L$ is the discrete Laplace-Beltrami operator and minimizes mean curvature. We construct $L$ once using the template mesh at the start of training.
% In keeping with a common practice across deformable model approaches
Following~\cite{BlanzVetter,Cashman,CSDM}, we also find it beneficial to regularize the deformations as it discourages arbitrarily large deformations and helps learn a meaningful mean shape. The corresponding energy term is expressed as $L_{\text{def}} = ||\Delta_V||_2$.  

%\paragraph{Updating the models.}

\subsubsection{Model Update.}
% \paragraph{Model Update.}
For iteratively refining the camera-multiplex, we use the summation of the silhouette and image reconstruction loss  $L_{\pi_k} = L_{\text{mask}, k} + L_{\text{pixel}, k}$ as the loss for each camera $\pi_k$ in the camera-multiplex. We optimize each camera to minimize $L_k$ every time the training instance is encountered during the training. For updating the shape and the texture, we use the resulting losses over the cameras as a distribution over the most likely camera pose in the camera-multiplex, and minimize the expected loss over all the cameras. Specifically, we compute the probability of $\pi_k$ being the optimal camera through a softmin function  $p_k = \frac{e^{-L_k/\sigma}}{\sum_j e^{-L_j/\sigma}}$ and train the shape and texture prediction modules with the final loss:
\begin{equation}
\label{eq:img-loss}
L_{\text{total}} =\sum_k p_k (L_{\text{mask}, k} + L_{\text{pixel}, k}) + L_\text{def} + L_\text{lap}.
\end{equation} 
In practice, the temperature $\sigma$ changes dynamically while computing $p_k$ by linearly normalizing $L_k$ to have a fixed range to standardize the peakiness of the probability distribution. We do not backpropagate through $p_k$. %as this observe that the optimized camera poses are more stable when we don't backpropagate through $w_i$. 
In summary, we iteratively refine the cameras in the multiplex against loss $L_{\pi_k}$ and update the parameters of $f$ through $L_\texttt{total}$ for every training sample. % \ak{maybe add something about who this is implemented in practice? complete a sentence like this (in experimental detail describe how you deal with scale augmentation, it could also come here): In practice, camera multiplex for each image $\mathcal{C}_i$ is a variable that is stored with each training image.. }

We implement the camera multiplex for each image $\mathcal{C}_i$ as a variable stored in a dictionary. Every time an image is encountered during training, the corresponding camera multiplex is fetched from this dictionary of variables and used as if it were an input to the rest of the training pipeline. Most modern deep learning frameworks such as PyTorch \cite{pytorch2019} support having such a dictionary of variables.

\subsubsection{Training a feed-forward camera predictor.}
When the training of $f$ converges, for each training image we select the optimal camera to be the camera that minimizes the silhouette and image reconstruction losses. We then train a new camera prediction module $g(I)$ in a supervised manner such that at inference time our model can predict all 3D shape, texture, and camera at the same time. 
%After training $f_\theta$ and optimizing camera multiplexes, we train a function $g_{\theta'}(I)$ to predict the camera pose at inference time. We model $g$ as a convolutional network and supervise it to predict the most probable camera pose $\pi^*$ in the camera multiplex $\pi_{1..K}$.

\subsubsection{Approach at test time.}
Given a novel image $I$ at test time, we can use the learnt modules $f$ and $g$ to predict the 3D shape, texture and camera-viewpoint of the object underlying image $I$. $f(I)$ predicts shape $V = \bar{V} + \Delta_V$ and texture $I^{uv}$ while $g(I)$ predicts the camera-viewpoint $\pi$. This is illustrated in Figure \ref{fig:overview}.

\section{Experiments}

%We evaluate our approach that learns to predict the 3D shape, texture, and camera from a single image without using any keypoint annotation during training time. We first describe our experimental setting. We e
In this section we provide quantitative and qualitative evaluation of our approach that learns to predict 3D shape, texture, and camera from a single image without using any keypoint annotations during training. We explore our approach on four object categories: birds, cars, motorcycles and shoes. %As no ground truth 3D shapes are available for any of these datasets, we will

\subsection{Experimental Detail}
\paragraph{Datasets.}
%%Dataset
We primarily use the CUB-200-2011 dataset \cite{Wah}, which has 6000 training and test images of 200 species of birds. In addition to this, we train and evaluate U-CMR on multiple categories: car, motorcycles from the Pascal3D+ dataset and shoes scraped from zappos.com. For CUB and Pascal3D, we use the same train-test splits as CMR \cite{cmrKanazawa18}. For the initial meshes for birds and cars, we use the 3D template meshes used by Kulkarni \etal \cite{kulkarni2019csm}. For others, we download freely available online models, homogenize to a sphere and simplify to reduce the number of vertices. We symmetrize all meshes to exploit bilateral symmetry. We compute masks for the zappos shoes dataset, which contains white background images, via simple threshold-based background subtraction and hole-filling.
% \sg{Add links to scraped datasets if possible}

\paragraph{Architecture.}
% Architecture / other detail
For all but texture, we use the same architecture as that of CMR \cite{cmrKanazawa18} and pass Resnet18 features into two modules - one each for predicting shape and texture.
% For most of the architecture, we follow that of CMR \cite{cmrKanazawa18}, which also recovers shape, texture, and camera pose from a single image using keypoint supervision during training. $f$ is a convolutional neural network with an encoder-decoder structure. It first extracts Resnet18 features from image $I$ into a latent feature map $z$ and then passes $z$ into two modules - one each for predicting shape and texture. 
The shape prediction module is a set of 2 fully connected layers with $\mathbb{R}^{3|V|}$ outputs that are reshaped into $\Delta_V$ following \cite{cmrKanazawa18}.
For the texture prediction, 
prior work predicted flow, where the final output is an offset that indicates where to sample pixels from. In this work 
we directly predict the pixel values of the UV image through a decoder. The texture head is a set of upconvolutional layers that takes the output of Resnet18 preserving the spatial dimensions. We find that this results in a more stable camera, as the decoder network is able to learn a spatial prior over the UV image. 
% We compare our version with both texture flow, and direct texture prediction. 
We use SoftRas \cite{liu2019soft} as our renderer. Please see the supplementary material for details and ablation studies. 

% As our focus is in analyzing the problem of learning a morphable shape model without keypoints, we focus our comparison against Category Specific Mesh Reconstruction (CMR) \cite{cmr}, which also learns a morphable shape model from image collections using both keypoints and silhouettes. Therefore we build on the exact same architecture as that of CMR, except that we do not use the parametric camera prediction module.

% \paragraph{Hardware and implementation details} We implemented out approach in PyTorch \cite{pytorch2019} and run our experiments on an Nvidia Titan RTX with $24$GB GPU memory on a machine with a $2.20$GHz $56$-core Intel(R) Xeon(R) Gold $5120$ CPU. We use the Adam \cite{kingma2014adam} for gradient-descent based optimization and use a batch-size of 12. It takes ~10 hours to run the optimization for the initial camera-multiplex initialization, 3 days for training shape+texture and another 2 hours for training feed-forward camera pose predictor. More details are shared in the appendix.

\paragraph{Camera-multiplex implementation.}
We use $K=40$ for camera-multiplex. We initialize the camera multiplex $\mathcal{C}$ for every image in the training set, to a set of $K$ points whose azimuth and elevation are uniformly spaced on the viewing sphere. For cars and motorcycles, we use $K=8$ cameras - all initialized to zero elevation. % to cover the entire azimuth-elevation space and are constrained to lie in a tessellation of the same.
We optimize each camera in the multiplex using the silhouette loss $L_{mask, k}$ before training shape and texture. To reduce compute time while training shape and texture, we reduce $K$ from 40 to 4 after 20 epochs by pruning the camera-multiplex and keeping the top $4$ cameras. %\ak{how do you deal with scale augmentation?}
Note that naive data augmentation that scales and/or translates the image without adjusting the camera-multiplex will result in the rendered shape being pixel-unaligned.
% Note that we can't use the camera multiplex for rendering directly when we're performing data augmentation that changes the size and position of the object in the image. 
We handle random crop and scale data augmentation during training by adjusting the scale and translation in the stored camera multiplex with a deterministic affine transformation before using it for rendering the shape.

\paragraph{Baselines.}
As no other approach predicts 3D shape, texture and pose without relying on keypoints or known camera or multi-view cues during training, as baseline we compare with ablations of our approach that do not use the camera-multiplex. This can be thought of CMR without keypoints, which simultaneously predicts shape, camera and texture and only supervises rendered silhouette and texture. We call this approach CMR-nokp and ensure that the experimental setup is comparable to U-CMR. Additionally, in the supplementary, we compare to two variants of CMR \cite{cmrKanazawa18} that have more supervision than our setting. 
% The first is the official CMR model (CMR-official) that was trained using additional keypoint losses and used NMR \cite{NMR} as it's differentiable renderer. The second is our implementation of CMR (CMR-ours) which is similar to U-CMR in its architecture for shape/texture, using Softras \cite{liu2019soft} for rendering and having the mean shape initialized to the same template mesh as U-CMR, but different from U-CMR in that it predicts a single camera pose and uses the ground-truth camera pose from SFM for supervision during training.  Unlike CMR-official, CMR-ours does not include vertex-keypoint reprojection loss. 
For camera prediction, we compare with CMR \cite{cmrKanazawa18} (which uses additional keypoint supervision), CSM \cite{kulkarni2019csm} and U-CMR without texture prediction (U-CMR-noTex).

\subsection{Qualitative Evaluation}
%\subsection{Problem with No Viewpoints and Keypoints}
% The task of predicting 3D shape, camera and texture from a single image is a fundamentally unconstrained problem. In particular, 3D shape and camera are the two sides of the same coin, where a grounding in one greatly reduces the amount of possible solution spaces. As such, previous works assume ground truth known viewpoints \cite{yan2016perspective,sitzmann2019srns} or solve for the viewpoints using the 2D semantic keypoint annotation as the correspondence \cite{cmrKanazawa18}. 
The problem when no keypoints and viewpoints are available is that there always exists a planar shape and texture that explains the image and silhouette for any arbitrary camera pose. We first demonstrate this point using CMR-nokp.
We observe that, as expected, CMR-nokp results in a degenerate solution shown in Figure~\ref{fig:degenerate-shape} where the recovered shape explains the image silhouette well, but when seen from a different viewpoint the shape is planar. 
\setlength{\columnsep}{10pt}
\setlength{\intextsep}{0pt}
\begin{wrapfigure}{r}{0.5\textwidth}
\begin{center}
% \begin{figure}[h]
  \centering
  \includegraphics[width=0.45\textwidth]{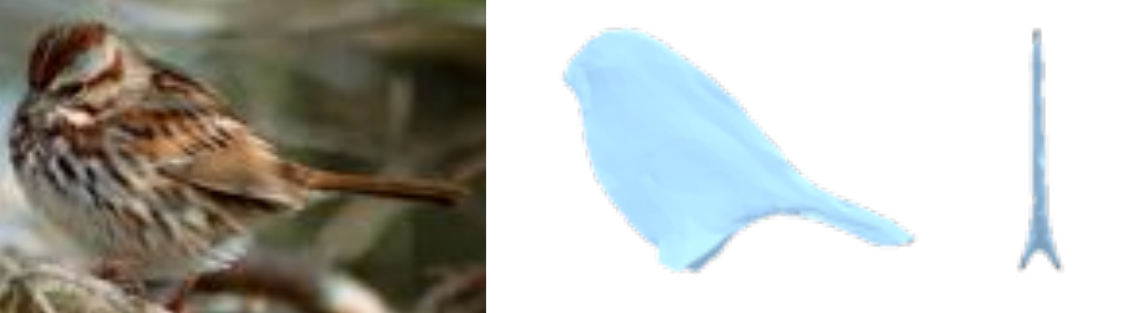}
%   \vspace{-1em}
% \end{figure}
\end{center}
\setlength{\intextsep}{0pt}
  \caption{\small{
%   {\bf Shape with no viewpoints and keypoints.} Degenerate shape when naively predicting the camera and shape from a single image without viewpoint and keypoint supervision. %This exemplifies the degenerate shapes we get by naively predicting camera and shape. 
%   Left is the input test image, the center is the shape from the camera viewpoint, which explains the silhouette well but an alternate view reveals that the model has learned a planar, flat bird shapes. %The image in the centre is a side-view of the predicted shape. The last image, which is a front-view of the shape, shows how the predicted shape is a degenerate flat surface with zero volume.
  {\bf CMR without keypoints.} 
  CMR-nokp, which directly predicts shape, texture, and camera without keypoint supervision or the proposed camera-multiplex, obtains degenerate solutions. The shape from predicted camera viewpoint (centre) explains the silhouette well but an alternate view (right) reveals that the model has learned to output a planar, flat bird shape.
  }}
    \label{fig:degenerate-shape}
\end{wrapfigure}

\begin{figure*}[h!]
  \centering
  \includegraphics[width=.99\textwidth]{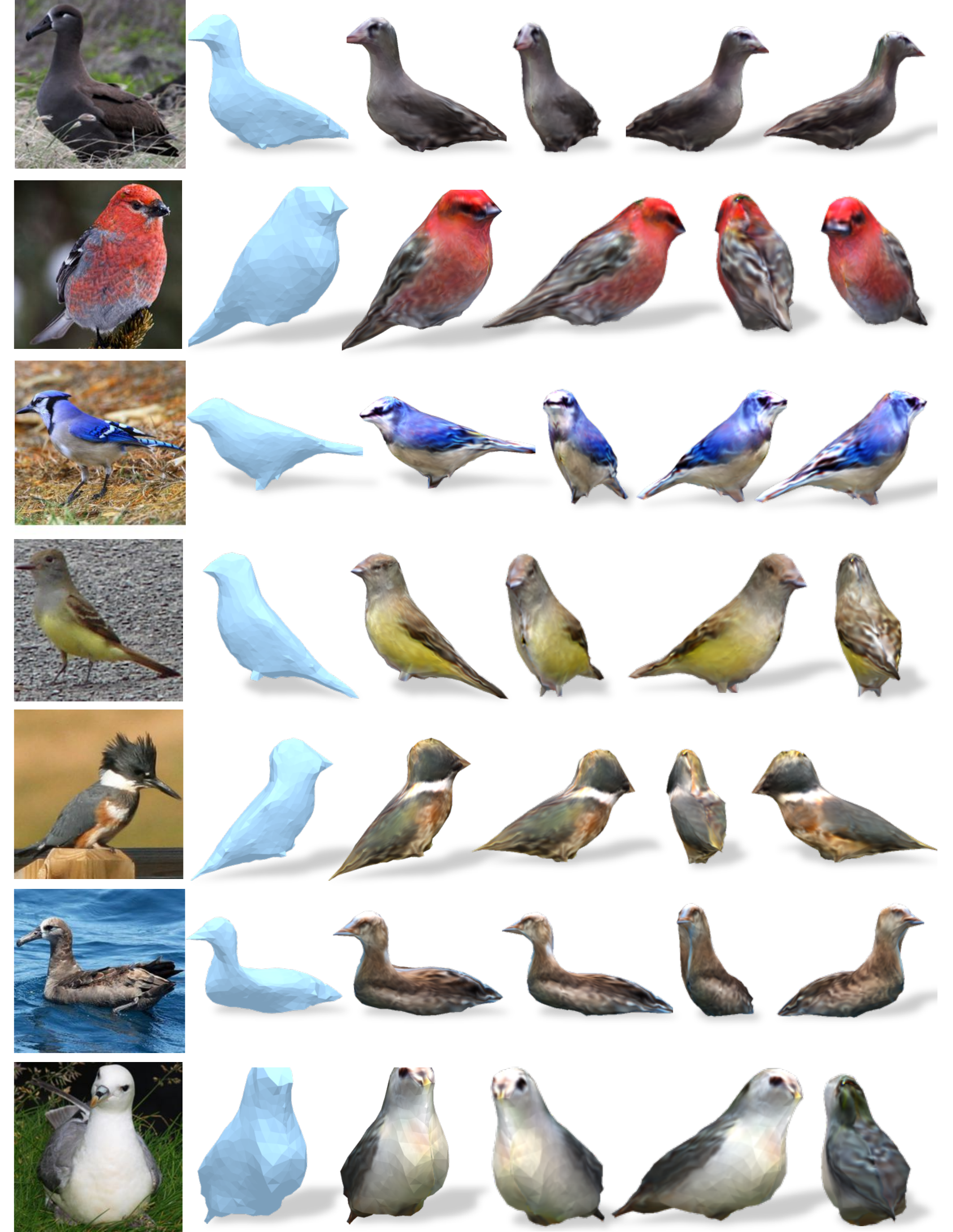}
  \caption{\small{{\bf Qualitative results.} For each input test image on the left, we show the predicted mesh, the textured mesh, and the textured mesh from multiple views. 
  %our shape with texture, different views. %The first column contains the input image and the second column renders the predicted shape from the predicted camera viewpoint. The third column adds the predicted texture. Columns 3-5 render the textured mesh from multiple views.
  }}
  %\vspace{-1em}
  \label{fig:big}
\end{figure*}

In Figure~\ref{fig:big}, we visualize U-CMR predictions on unseen images from the CUB test set. 
Our approach, despite not using any viewpoint or keypoint supervision is able to recover a full, plausible 3D shape of the birds and learns to predict their texture from a single image. Our approach captures various types of bird shapes, including shapes of water birds and songbirds. We are able to recover sharp long tails and some protrusion of legs and beaks. 
Please see supplementary for more results of random samples from test set and comparisons to CMR.
% Observe that CMR-official is not as accurate as CMR-ours in capturing the shape and texture of the underlying bird but has pointier beaks and feet because of the keypoint reprojection loss it uses. The figures show that U-CMR shapes are qualitatively very similar to CMR-ours, hence exemplifying our assertion that U-CMR's camera-multiplex optimization alleviates the need for ground-truth camera pose supervision for most cases.

\begin{figure}[h]
  \centering
%   \vspace{1em}
  \includegraphics[width=.9\textwidth]{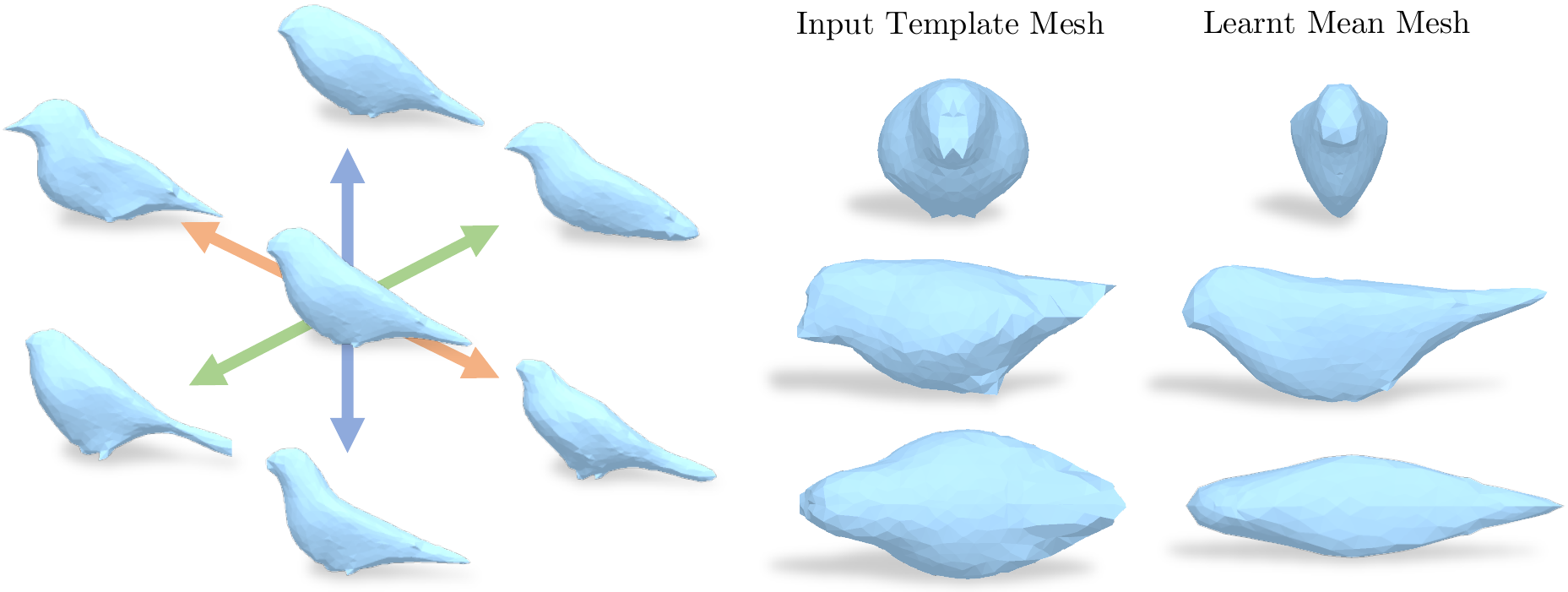}
  \caption{\small{{\bf Learned Shape.} On the left, we visualize the space of learned shapes by running PCA. We see that the model has learned to output changes in body type, head shape, and tail types. On the right, we compare the template shape to the final learnt mean mesh. See text for discussion.
  }}
%   \vspace{-2em}
  \label{fig:pca}
\end{figure}

We further analyze the shape space that we learn in Figure \ref{fig:pca}, where we run principal component analysis on all the shapes obtained on the train set. We find directions that capture changes in the body type, the head shapes, and the tail shapes. In Figure \ref{fig:pca}, we also show that the final mean shape deviates significantly from the template mesh it was initialized to, by becoming thinner and developing a more prominent tail. Please see the supplemental for more results.

% \setlength{\columnsep}{10pt}
% \setlength{\intextsep}{0pt}
% \begin{wrapfigure}{r}{0.5\textwidth}
%   \begin{center}
% \includegraphics[width=.5\textwidth]{figures/camera-dist.png}
%   \end{center}
%   %\vspace{-1em}
%   \label{fig:camera-dist}
% \setlength{\intextsep}{0pt}
% \end{wrapfigure}

\setlength{\tabcolsep}{4pt}
\begin{center}
\begin{table}[t]
\centering
\begin{tabular}{c|cccc}
\toprule%\hline
\multicolumn{1}{c|}{} & \multicolumn{1}{c}{\multirow{2}{*}{\begin{tabular}[c]{@{}c@{}}Rotation\\ Error $\downarrow$ \end{tabular}}} & \multicolumn{1}{c}{\multirow{2}{*}{\begin{tabular}[c]{@{}c@{}}Entropy\\ (nats) $\uparrow$ \end{tabular}}} & \multicolumn{2}{c}{\begin{tabular}[c]{@{}c@{}}Wasserstein Dist $\downarrow$ \end{tabular}} \\
\multicolumn{1}{c|}{} & \multicolumn{1}{c}{}                                        & \multicolumn{1}{c}{}                                                                          & \multicolumn{1}{c}{Azimuth}                  & \multicolumn{1}{c}{Elevation}                 \\
\hline
GT  & - & 7.44  & - & -   \\
CMR \cite{cmrKanazawa18}    & 22.94 \degree  & 7.25  & 6.03 \degree  & 4.33 \degree  \\
\hline
CMR-nokp                    & 87.52 \degree  & 5.73  & 64.66 \degree & 12.39 \degree \\
CSM  \cite{kulkarni2019csm} & 61.93 \degree  & 5.83  & 27.34 \degree & 11.28 \degree \\
U-CMR (noTex)               & 61.82 \degree  & \textbf{7.36} & 16.08 \degree & 7.90 \degree \\
U-CMR         & \textbf{45.52 \degree} & 7.26 & \textbf{8.66 \degree } & \textbf{6.50 \degree  } \\
\bottomrule
\end{tabular}
\caption{\small{{\bf Quantitative evaluation of camera pose predictions on the test dataset.} We plot rotation error as the geodesic distance from the ground-truth, the entropy of the azimuth-elevation distribution and the wasserstein distance of marginal Az/El \wrt ground-truth. U-CMR outperforms all methods in the absence of keypoints.}
}
\label{table:camera}
% \vspace{-2em}
\end{table}
\end{center}
\setlength{\tabcolsep}{1.4pt}

\begin{figure}[t]
%   \vspace{1em}
  \centering
  \includegraphics[width=\textwidth]{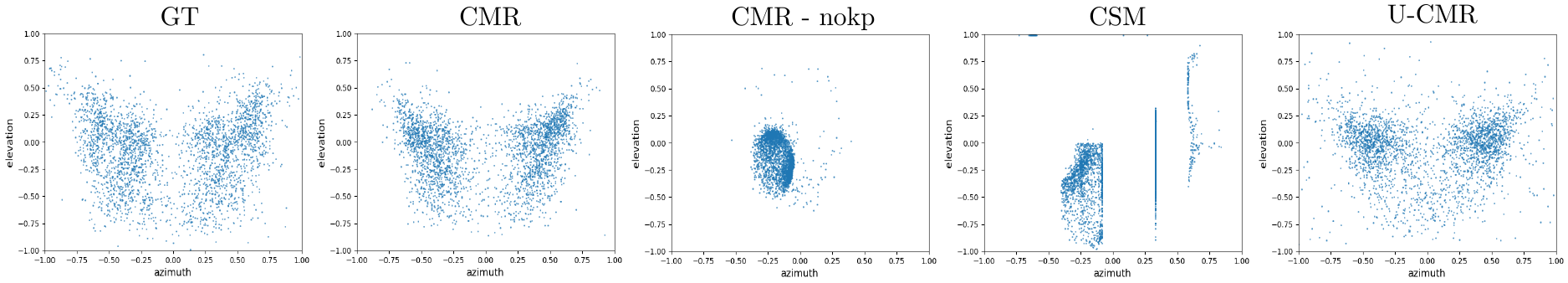}
  \caption{\small{{\bf Camera Pose Distributions on CUB Test.} We show azimuth-elevation scatter plots over the entire CUB test set for different approaches. From left to right, we show (i) pseudo ground-truth cameras computed
  via running SfM on keypoints,%using SfM on the set of all keypoints, 
  (ii) the cameras predicted by CMR which uses the SfM cameras as supervision, (iii) CMR without viewpoint and keypoint supervision (CMR-nokp), (iv) CSM \cite{kulkarni2019csm} and (v) our approach U-CMR. The last three approaches are weakly-supervised and do not use any keypoint annotation. Notice how the camera pose collapses in CMR-nokp and CSM, while U-CMR with camera-multiplex is able to obtain distribution similar to the ground truth cameras.}}
%   \vspace{-1em}
  \label{fig:camera-dist}
\end{figure}

\subsection{Quantitative Evaluation}
We conduct quantitative evaluation on the camera poses obtained from our approach, since there are no 3D ground truth shapes on this dataset.
%We further evaluate the camera poses from our approach quantitatively. Since ground truth shape is not available for the CUB dataset, we do not evaluate our shapes quantitatively. 
For camera evaluation, we compare our approach to CSM \cite{kulkarni2019csm}, which learns to output dense correspondences of the image against a template 3D shape as well as the camera poses from image collections without keypoints. Note that they do not learn to predict 3D shapes. %CSM attempts to overcome the camera collapse issue via having a multi-pose camera prediction head that outputs 8 possible camera poses with entropy and diversity constraints that encourages a diverse camera prediction.
We used the same 3D template mesh as CSM and therefore are comparable to CSM.
We evaluate cameras from different approaches on metrics measuring their accuracy and collapse. We compare our predicted cameras to the pseudo ground-truth cameras computed in CMR \cite{cmrKanazawa18} using SFM on keypoints. 
% Recall that we're using a weak perspective camera parametrized by scale $\textbf{s} \in \mathbb{R}$, translation $\textbf{t} \in\mathbb{R}^2$ and rotation $\textbf{R}$ (captured as euler angles azimuth, elevation, cyclo-rotation $[\textit{az,el,cr}] \in\mathbb{R}^3$). 
For evaluating accuracy, we compute the rotation error $\text{err}_{R} = \arccos\left(\frac{\text{Tr}(\tilde{R}^TR^*) - 1}{2}\right)$ as the geodesic distance between the predicted camera rotation $\tilde{R}$ and the pseudo ground-truth camera rotation $R^*$. We report the average rotation error (in degrees) over the entire test dataset. To measure collapse, we analyze the azimuth-rotation distribution and report (i) its entropy (in nats) and (ii) it's Wasserstein distance to the pseudo ground truth azimuth-elevation distribution. Because of computational ease, we only report the Wasserstein distance on the azimuth and elevation marginals. We primarily focus on azimuth and rotation because changes in scale, translation and cyclo-rotation of a camera only warp the image in 2D and don't constitute a ``novel viewpoint". 

Table~\ref{table:camera} reports the numbers on all metrics for supervised (CMR) and weakly-supervised (CMR-nokp, CSM, U-CMR) methods. Observe that all the weakly-supervised baselines incur significant camera pose collapse - as can be seen by the entropy of their distributions. In contrast, U-CMR, despite being weakly-supervised, achieves an entropy that is slightly better than CMR - the supervised baseline. We automatically learn a camera distribution that is almost as close to the ground-truth distribution (in Wasserstein distance) as the supervised baseline (CMR). U-CMR is more accurate than CMR-nokp and CSM and achieves an average rotation error at least 15 degrees better than them. This table also suggests that the texture loss helps with refining camera poses to make them more accurate as U-CMR (noTex) is well-distributed with a very high entropy but is not as accurate as U-CMR. 

Figure~\ref{fig:camera-dist} visualizes the azimuth-elevation distributions of different approaches. This figure illustrates that while CSM prevents an extreme mode collapse of the camera, their camera pose distribution still collapses into certain modes. For making this figure, we employ CSM's public model, trained on the same CUB dataset with a fixed template shape. This empirical evidence exemplifies the fact that U-CMR's weakly-supervised camera-multiplex optimization approach learns a camera pose distribution that's much better than other weakly-supervised baselines and almost as good as supervised methods.

% Next, we conduct several quantitative evaluation on that indirectly measures the quality of the shape and the camera learned by our model. We evaluate camera accuracy on training and the test set, and also evaluate on the mask reprojection accuracy, which requires good shape as well as camera prediction on the test set. 

% \begin{figure*}[h]
%   \centering
%   \includegraphics[width=\textwidth,height=.2\textheight,draft]{figures/traincamera.pdf}
%   \caption{\small{{\bf Visualization of estimated camera distribution on training set.} .}}
%   %\vspace{-1em}
%   \label{fig:traincamera}
% \end{figure*}

\begin{figure*}[h!]
  \centering
%   \vspace{1em}
  \includegraphics[width=\textwidth]{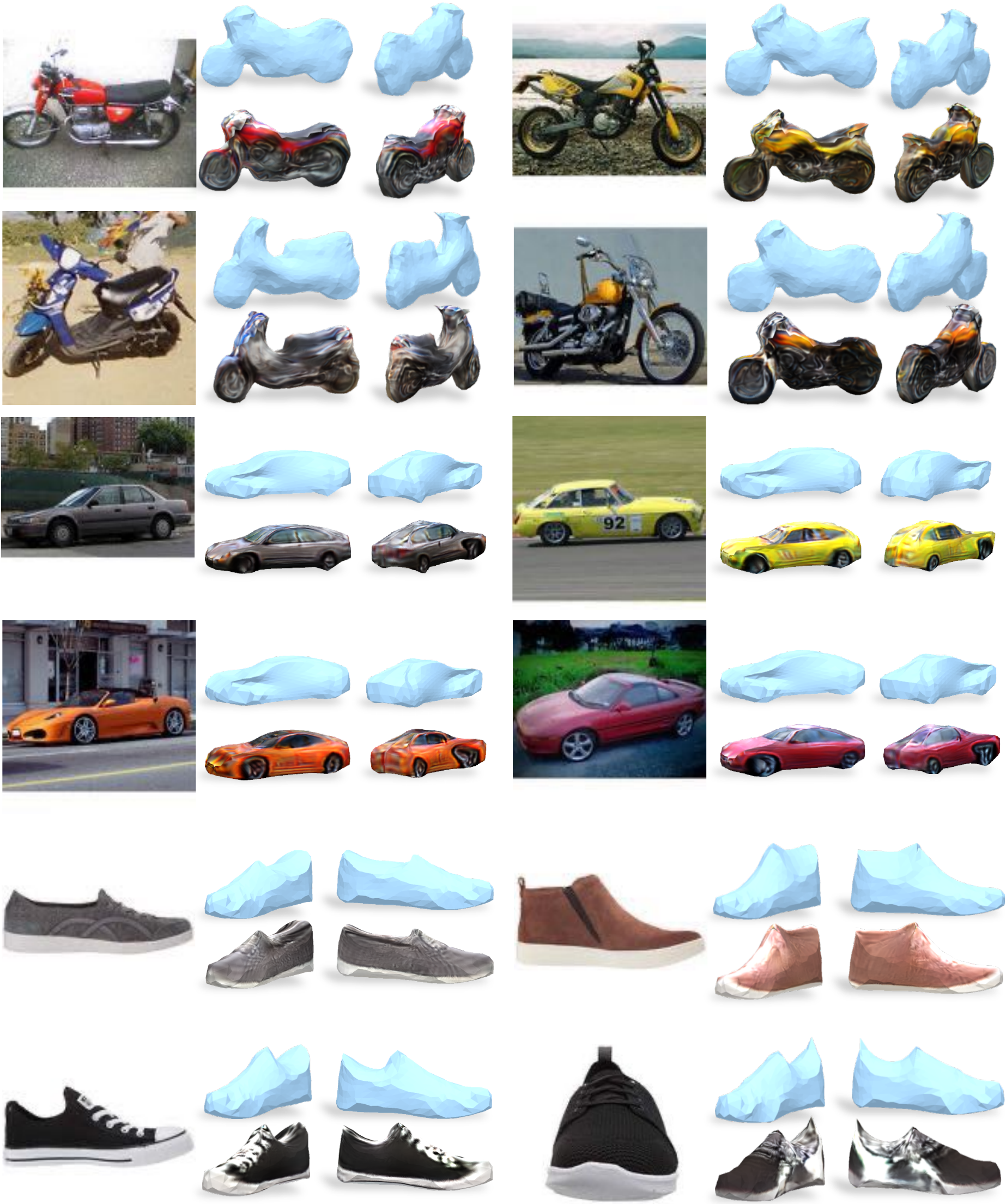}
  \caption{\small{{\bf Qualitative results on cars, motorcycles and shoes.}
  %We show predictions of our approach on images from the test set for Pascal3D cars and motorcycles and Zappos shoes.
  For each image, we show the predicted 3D shape and texture from two viewpoints.
  %For image input on the top left, we show (in order) the predicted 3D shape and texture from two viewpoints.
  }}
%   \vspace{-1em}
  \label{fig:big-other}
\end{figure*}

\subsection{Evaluations on other categories}
While our primary evaluation is on the CUB Birds dataset, we also run our approach on  cars and motorcycles from the Pascal3D+ dataset, and the shoe images we scraped from zappos.com. We show qualitative visualizations of predicted shape, texture and camera for all categories. For Pascal3D cars, we also compare IoU of predicted shapes to CMR, previous deformable model fitting-based~\cite{CSDM} and volumetric prediction ~\cite{drcTulsiani17} methods. All three of these approaches leverage segmentation masks and cameras and keypoints to learn 3D shape inference.

Figure~\ref{fig:big-other} shows qualitative results on selected images from their respective test set. U-CMR learns accurate camera poses and diverse yet plausible shapes for cars and motorcycles. For shoes, U-CMR shapes are reasonable but not as diverse because of biases in the underlying dataset. 
We observe some artifacts where the sides of the cars have concave indentations and some parts of the shoes are tapered and pointy. These issues stem from using weak-perspective projections and limitations of the regularization, which is not spatially adaptive. Please see the supplemental for each category's PCA visualizations and the initial template.

We report the mean IoU on the test set in Table~\ref{table:p3d} and observe that U-CMR performs comparably to alternate methods that require more supervision.

\begin{table}
    % \vspace{1em}
    \centering
    \begin{tabular}{c | c c c c}
    \toprule
    Category & CSDM~\cite{CSDM} & DRC~\cite{drcTulsiani17} &  CMR~\cite{cmrKanazawa18}   & U-CMR \\ \midrule
    Car &   0.60  & 0.67 & 0.640 & 0.646 \\ 
    % Motorbike &   0.00  & 0.00 & 0.000 & 0.000 \\ 
    % Aeroplane &   0.00  & 0.00 & 0.000 & 0.000 \\ 
    \bottomrule
    \end{tabular}
      \caption{\small {\bf Reconstruction evaluation using PASCAL 3D+.} We report the mean intersection over union (IoU) on PASCAL 3D+ to benchmark the obtained 3D reconstructions (higher is better). We compare to CMR \cite{cmrKanazawa18}, a deformable model fitting approach (CSDM \cite{CSDM}) and a volumetric prediction approach (DRC \cite{drcTulsiani17}). CSDM and DRC use image collection supervision in addition to keypoints/cameras.}
      \label{table:p3d}
    %   \vspace{-3em}
\end{table}

\subsection{Limitations}
\begin{figure}[h!]
  \centering
%   \vspace{1em}
  \includegraphics[width=0.95\textwidth]{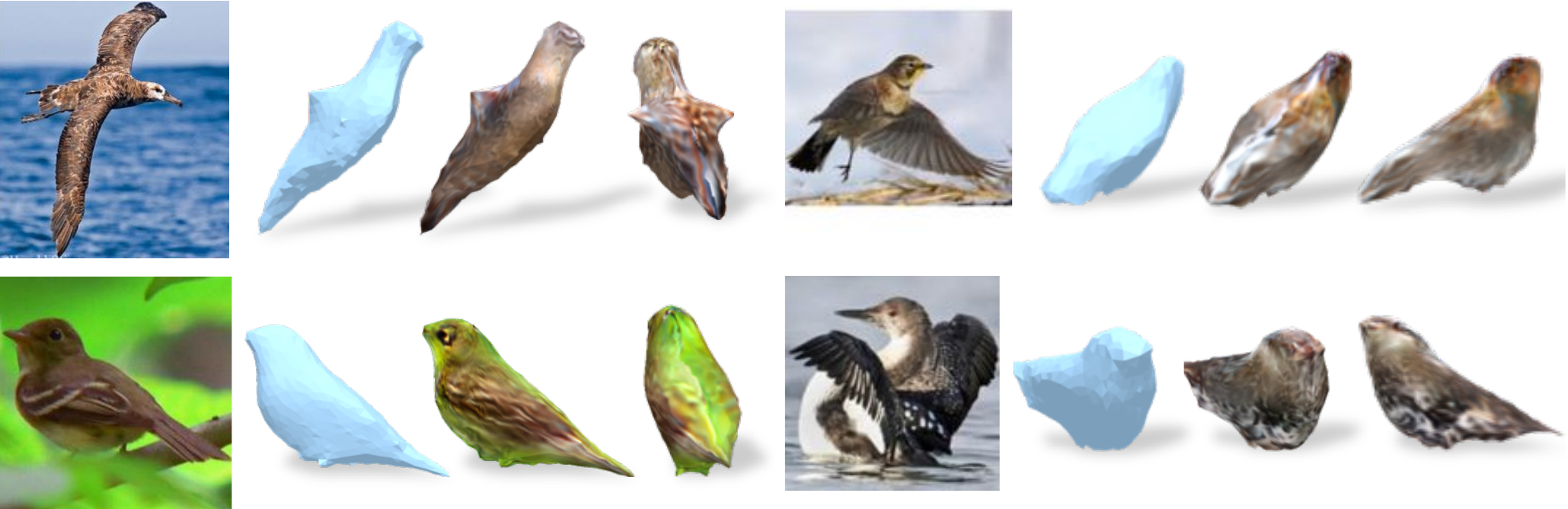}
  \caption{\small{ {\bf Failure Modes.} The columns, from left to right, show the input image, the predicted shape and texture from the predicted camera, and finally a different view of the textured mesh. See the text for discussion.
  %The bottom left
%   The two figures in the top row show how our shape prediction fails when the bird in the image is flying, or has it's wings open because the shape is very different from the mean shape. In these cases, the predicted shape looks like a bird but usually doesn't capture the instance in the image.
%   The example on the bottom left shows how background colours sometimes leak into the predicted texture. The other three examples show typical failure in shape and pose prediction when the bird articulates too much.
% The example in the bottom right shows an articulated bird with it's head twisted up. Since we do not not model articulation, pose and shape prediction struggles in these cases.
  }}
%   \vspace{-1em}
  \label{fig:limitation}
\end{figure}

% \paragraph{Limitations.} 
While U-CMR shows promising results in the direction of weakly supervised 3D shape understanding, it has some limitations. Foremost limitation is that we do not model articulation and expect to fail in cases with significant articulation. In Figure~\ref{fig:limitation}, we demonstrate various modes of failure for shape, texture and camera-viewpoint prediction. Our approach struggles when the bird shape is significantly different from the template mesh and undergoes large articulation, such as the case with flying birds. It is challenging to identify correct camera poses when there's a large deformation like this without keypoints. The data imbalance between flying and not flying birds also exacerbates this problem. The two examples in the top row of the figure show how our shape prediction fails when the bird in the image is flying, or has it's wings open. The example in the bottom right shows an articulated bird with it's head twisted back. Due to the lack of an articulation model, these failure cases are expected. We also fail at predicting good texture sometimes - especially for parts of the object that are not visible. The example on the bottom left of Figure~\ref{fig:limitation} and bottom right of Figure~\ref{fig:big-other} shows how background colours may leak into the predicted texture.

% \begin{figure*}[h]
%   \centering
%   \vspace{-2em}
%   \includegraphics[width=0.8\textwidth]{suppl_figures/FIG random car.pdf}
%   \caption{\small{{\bf Qualitative results on Pascal3D cars.} We show predictions of our approach on images from the test set. For image input on the top left, we show (in order) the predicted 3D shape from the predicted camera, the predicted texture from the predicted camera and 2 novel views of the predicted 3D mesh.
%   }}
%   \vspace{-2em}
%   \label{fig:big-car}
% \end{figure*}

% \begin{figure*}[h]
%   \centering
%   \vspace{-2em}
%   \includegraphics[width=0.8\textwidth, draft]{figures/limit.pdf}
%   \caption{\small{{\bf Qualitative results on Pascal3D X.} We show predictions of our approach on images from the test set. For image input on the top left, we show (in order) the predicted 3D shape from the predicted camera, the predicted texture from the predicted camera and 2 novel views of the predicted 3D mesh.
%   }}
%   \vspace{-2em}
%   \label{fig:big-tv}
% \end{figure*}

% We use the template 3D shapes used by Kulkarni et. al. \cite{csm}

\section{Conclusion}
In this work, we present a learning framework that can decompose an image of a deformable object into its 3D shape, texture, and camera viewpoint. In order to solve this highly under-constrained problem, we propose a representation for maintaining a distribution over possible camera viewpoints called camera-multiplex. This allows the model to maintain a possible set of camera hypothesis, avoiding the learning process from getting stuck in a bad local minima. We show our approach on four categories, where we show that U-CMR can recover reasonable 3D shape and texture without viewpoints and keypoints. %obtain state of the art camera prediction accuracy under the setting of no viewpoint and no keypoints. 

~\\
\noindent\textbf{Acknowledgements.} We thank Jasmine Collins for scraping the zappos shoes dataset and members of the BAIR community for helpful discussions. This work was supported in-part by eBay, Stanford MURI and the DARPA MCS program.

% \clearpage
\bibliographystyle{splncs04}

% \bibliography{egbib}

\begin{thebibliography}{10}
\providecommand{\url}[1]{\texttt{#1}}
\providecommand{\urlprefix}{URL }
\providecommand{\doi}[1]{https://doi.org/#1}

\bibitem{BlanzVetter}
Blanz, V., Vetter, T.: A morphable model for the synthesis of 3d faces. In:
  SIGGRAPH (1999)

\bibitem{blender}
{Blender Online Community}: Blender - a 3D modelling and rendering package.
  Blender Institute, Amsterdam (2019), \url{http://www.blender.org}

\bibitem{Cashman}
Cashman, T.J., Fitzgibbon, A.W.: What shape are dolphins? building 3{D}
  morphable models from 2{D} images. TPAMI  (2013)

\bibitem{chen2016synthesizing}
Chen, W., Wang, H., Li, Y., Su, H., Wang, Z., Tu, C., Lischinski, D., Cohen-Or,
  D., Chen, B.: Synthesizing training images for boosting human 3d pose
  estimation. In: 3DV (2016)

\bibitem{choy20163d}
Choy, C.B., Xu, D., Gwak, J., Chen, K., Savarese, S.: 3d-r2n2: A unified
  approach for single and multi-view 3d object reconstruction. In: ECCV (2016)

\bibitem{dai2014simple}
Dai, Y., Li, H., He, M.: A simple prior-free method for non-rigid
  structure-from-motion factorization. IJCV  (2014)

\bibitem{psgn}
Fan, H., Su, H., Guibas, L.J.: A point set generation network for 3d object
  reconstruction from a single image. In: CVPR (2017)

\bibitem{gadelha20173d}
Gadelha, M., Maji, S., Wang, R.: 3d shape induction from 2d views of multiple
  objects. In: 3DV (2017)

\bibitem{girdhar16b}
Girdhar, R., Fouhey, D., Rodriguez, M., Gupta, A.: Learning a predictable and
  generative vector representation for objects. In: ECCV (2016)

\bibitem{meshrcnn}
Gkioxari, G., Malik, J., Johnson, J.: Mesh r-cnn. In: ICCV (2019)

\bibitem{hughes2014computer}
Hughes, J.F., Foley, J.D.: Computer graphics: principles and practice. Pearson
  Education (2014)

\bibitem{insafutdinov2018unsupervised}
Insafutdinov, E., Dosovitskiy, A.: Unsupervised learning of shape and pose with
  differentiable point clouds. In: NeurIPS (2018)

\bibitem{jakab2018unsupervised}
Jakab, T., Gupta, A., Bilen, H., Vedaldi, A.: Unsupervised learning of object
  landmarks through conditional image generation. In: NeurIPS (2018)

\bibitem{kanazawa2016learning}
Kanazawa, A., Kovalsky, S., Basri, R., Jacobs, D.: Learning 3d deformation of
  animals from 2d images. In: Computer Graphics Forum. Wiley Online Library
  (2016)

\bibitem{cmrKanazawa18}
Kanazawa, A., Tulsiani, S., Efros, A.A., Malik, J.: Learning category-specific
  mesh reconstruction from image collections. In: ECCV (2018)

\bibitem{lsmKarHM2017}
Kar, A., H\"ane, C., Malik, J.: Learning a multi-view stereo machine. In:
  NeurIPS (2017)

\bibitem{CSDM}
Kar, A., Tulsiani, S., Carreira, J., Malik, J.: Category-specific object
  reconstruction from a single image. In: CVPR (2015)

\bibitem{NMR}
Kato, H., Ushiku, Y., Harada, T.: Neural 3d mesh renderer. In: CVPR (2018)

\bibitem{kulkarni2019csm}
Kulkarni, N., Gupta, A., Tulsiani, S.: Canonical surface mapping via geometric
  cycle consistency. In: ICCV (2019)

\bibitem{liu2019soft}
Liu, S., Li, T., Chen, W., Li, H.: Soft rasterizer: A differentiable renderer
  for image-based 3d reasoning. In: ICCV (2019)

\bibitem{loper2014opendr}
Loper, M.M., Black, M.J.: Opendr: An approximate differentiable renderer. In:
  ECCV (2014)

\bibitem{novotny2019c3dpo}
Novotny, D., Ravi, N., Graham, B., Neverova, N., Vedaldi, A.: C3dpo: Canonical
  3d pose networks for non-rigid structure from motion. In: ICCV (2019)

\bibitem{park2019deepsdf}
Park, J.J., Florence, P., Straub, J., Newcombe, R., Lovegrove, S.: Deepsdf:
  Learning continuous signed distance functions for shape representation. In:
  CVPR (2019)

\bibitem{pytorch2019}
Paszke, A., Gross, S., Massa, F., Lerer, A., Bradbury, J., Chanan, G., Killeen,
  T., Lin, Z., Gimelshein, N., Antiga, L., Desmaison, A., Kopf, A., Yang, E.,
  DeVito, Z., Raison, M., Tejani, A., Chilamkurthy, S., Steiner, B., Fang, L.,
  Bai, J., Chintala, S.: Pytorch: An imperative style, high-performance deep
  learning library. In: NeurIPS (2019)

\bibitem{pinkall1993computing}
Pinkall, U., Polthier, K.: Computing discrete minimal surfaces and their
  conjugates. Experimental mathematics  (1993)

\bibitem{shu2018deforming}
Shu, Z., Sahasrabudhe, M., Alp~Guler, R., Samaras, D., Paragios, N., Kokkinos,
  I.: Deforming autoencoders: Unsupervised disentangling of shape and
  appearance. In: ECCV (2018)

\bibitem{song2017semantic}
Song, S., Yu, F., Zeng, A., Chang, A.X., Savva, M., Funkhouser, T.: Semantic
  scene completion from a single depth image. In: CVPR (2017)

\bibitem{thewlis2017unsupervised}
Thewlis, J., Bilen, H., Vedaldi, A.: Unsupervised learning of object frames by
  dense equivariant image labelling. In: NeurIPS (2017)

\bibitem{torresani2008nonrigid}
Torresani, L., Hertzmann, A., Bregler, C.: Nonrigid structure-from-motion:
  Estimating shape and motion with hierarchical priors. TPAMI  (2008)

\bibitem{mvcTulsiani18}
Tulsiani, S., Efros, A.A., Malik, J.: Multi-view consistency as supervisory
  signal for learning shape and pose prediction. In: CVPR (2018)

\bibitem{factored3dTulsiani17}
Tulsiani, S., Gupta, S., Fouhey, D., Efros, A.A., Malik, J.: Factoring shape,
  pose, and layout from the 2d image of a 3d scene. In: CVPR (2018)

\bibitem{drcTulsiani17}
Tulsiani, S., Zhou, T., Efros, A.A., Malik, J.: Multi-view supervision for
  single-view reconstruction via differentiable ray consistency. In: CVPR
  (2017)

\bibitem{varol17_surreal}
Varol, G., Romero, J., Martin, X., Mahmood, N., Black, M.J., Laptev, I.,
  Schmid, C.: Learning from synthetic humans. In: CVPR (2017)

\bibitem{Vincente}
Vicente, S., Carreira, J., Agapito, L., Batista, J.: Reconstructing pascal voc.
  In: CVPR (2014)

\bibitem{Wah}
Wah, C., Branson, S., Welinder, P., Perona, P., Belongie, S.: {The Caltech-UCSD
  Birds-200-2011 Dataset}. Tech. Rep. CNS-TR-2011-001, California Institute of
  Technology (2011)

\bibitem{wu2019}
Wu, S., Rupprecht, C., Vedaldi, A.: Unsupervised learning of probably symmetric
  deformable 3d objects from images in the wild. arXiv preprint
  arXiv:1911.11130  (2019)

\bibitem{wu20153d}
Wu, Z., Song, S., Khosla, A., Yu, F., Zhang, L., Tang, X., Xiao, J.: 3d
  shapenets: A deep representation for volumetric shapes. In: CVPR (2015)

\bibitem{xiang2014beyond}
Xiang, Y., Mottaghi, R., Savarese, S.: Beyond pascal: A benchmark for 3d object
  detection in the wild. In: WACV (2014)

\bibitem{yan2016perspective}
Yan, X., Yang, J., Yumer, E., Guo, Y., Lee, H.: Perspective transformer nets:
  Learning single-view 3d object reconstruction without 3d supervision. In:
  NeurIPS (2016)

\bibitem{zhang2018perceptual}
Zhang, R., Isola, P., Efros, A.A., Shechtman, E., Wang, O.: The unreasonable
  effectiveness of deep networks as a perceptual metric. In: CVPR (2018)

\bibitem{Zuffi19Safari}
Zuffi, S., Kanazawa, A., Berger-Wolf, T., Black, M.J.: Three-d safari: Learning
  to estimate zebra pose, shape, and texture from images "in the wild". In:
  ICCV (2019)

\end{thebibliography}

\clearpage
\section*{Appendix}
\subsection*{Overview}
In this supplementary material, we present additional results including visualization of the shape, texture and camera-multiplex, comparisons to CMR \cite{cmrKanazawa18}, qualitative results on random test samples, an ablation study on texture prediction model, and additional details on the network architecture.

\section{More Results}

% \subsection{Camera-multiplex visualization over time.}
% Figure~\ref{fig:camera-dist-time} shows how the azimuth-elevation distribution of camera poses in the camera-multiplex (over the entire training dataset) changes as training progresses. Observe that the distribution changes rapidly initially and results in a final distribution that is very different from the initial distribution.

% \begin{figure}[h]
%   \centering
%   \includegraphics[width=\textwidth]{suppl_figures/cameras over time.png}
%   \caption{{{\bf Change in Camera Pose Distributions during training.} We show an azimuth-elevation scatter plot of $K=40$ camera poses in all the camera multiplex of the training dataset at different points during training. Points corresponding to less probable cameras have a lower alpha value and are more transparent. Starting from top left, we have the camera poses after the camera-multiplex initialization, after 1 training epoch, after 20 training epochs and the final optimized camera poses. The number of camera poses in each multiplex ($K$) is pruned down from $40$ to $4$ after epoch 20.}}
%   %\vspace{-1em}
%   \label{fig:camera-dist-time}
% \end{figure}

\subsubsection{PCA in Texture space}
\begin{figure}[h]
  \centering
%   \vspace{1em}
  \includegraphics[width=\textwidth]{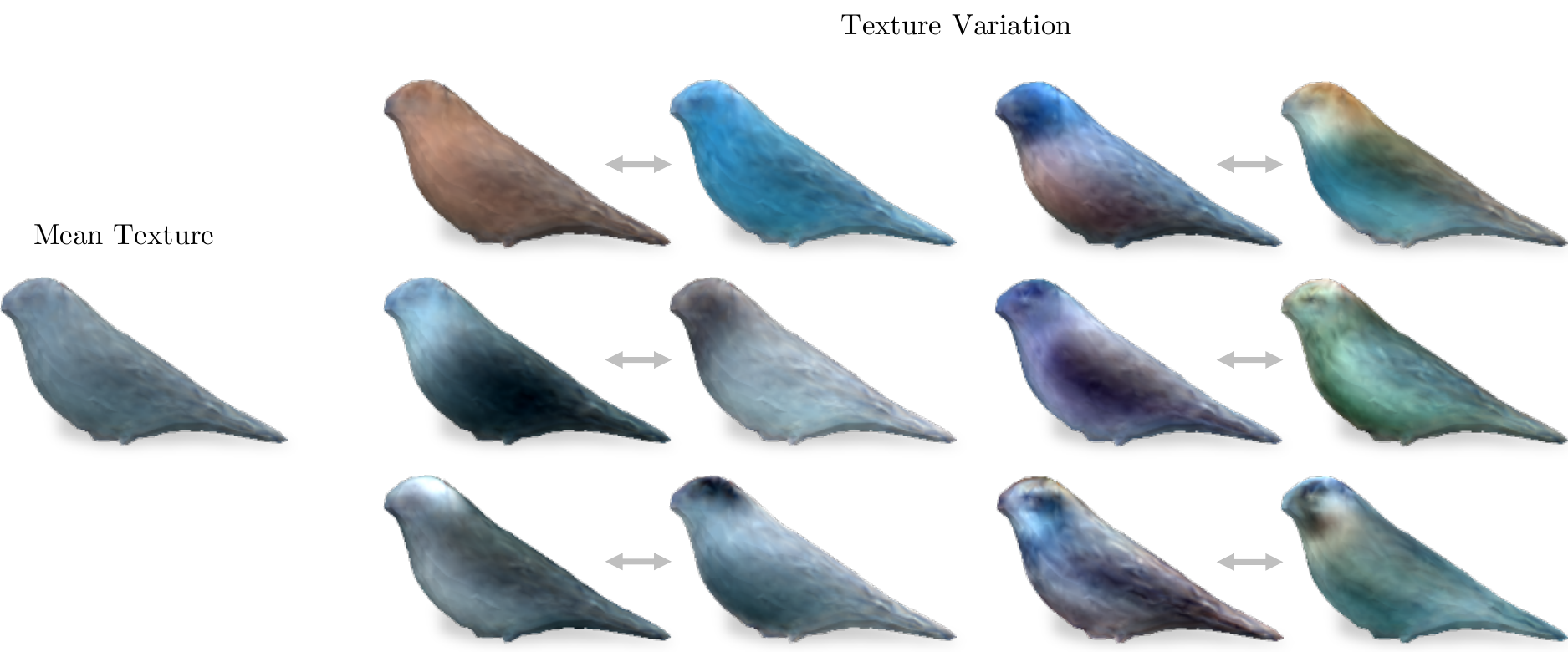}
  \caption{\small{{\bf PCA in UV-Texture space.}
  We visualize the learnt texture space by running PCA over all predicted uv-texture maps on the test dataset, and rendering axes of variation on the learnt mean bird shape. See text for discussion. }}
%   \vspace{-1em}
  \label{fig:pca_tex}
\end{figure}

In Figure~\ref{fig:pca_tex}, we visualize the learnt texture space by running PCA on predicted uv-textures across the entire test dataset. On the left, the mean texture is a rather dull gray colour, as expected. On the right, we visualize axes of variation. In the first column, we see low-frequency variations in overall colour, head and belly of the bird. In the second column, we see slightly higher-frequency variations that assign different colors to different parts (head, back, belly and wings) of the bird. We can even recognize eyes and beak in the last two rows on the right.

\subsubsection{Learnt shape space for other categories}
\begin{figure}[!h]
  \centering
%   \vspace{1em}
  \includegraphics[width=0.92\textwidth]{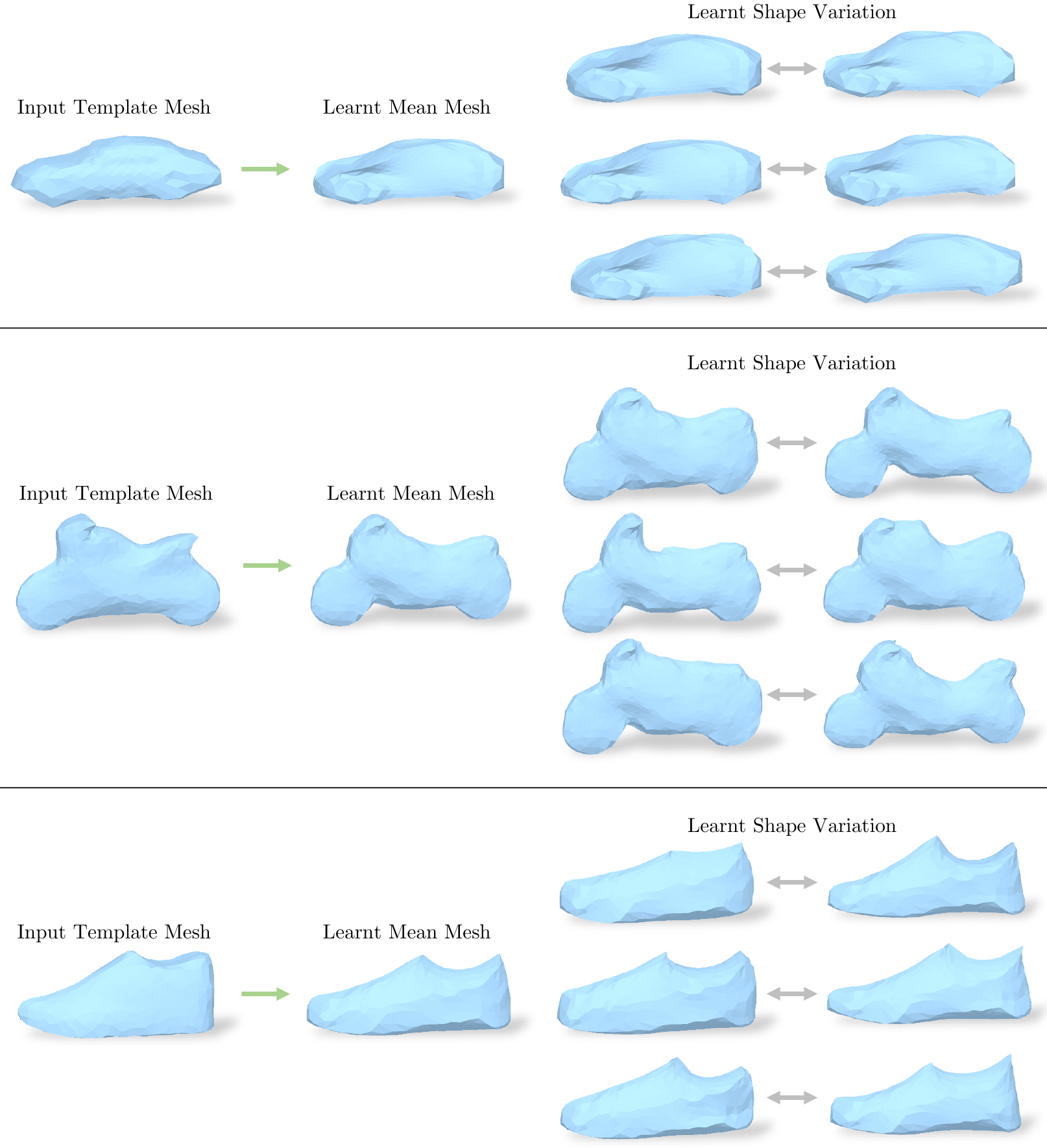}
  \caption{\small{{\bf Learned Shape on other categories.} On the left, we compare the template shape to the final learnt mean mesh. On the right, we visualize the space of learned shapes by running PCA on all shapes and show three axis of deformations.
  }}
  %\vspace{-1em}
  \label{fig:other-pca}
\end{figure}

In Figure~\ref{fig:other-pca}, we compare the input template mesh to the learnt shape space for car, motorbike and shoe categories.
Observe, on the left, that the input template and learnt mean shape differ substantially for each of these categories - some more than others. For example, the front tire of the motorbike becomes more significantly more prominent, the back of the car becomes more rounded and the shoe becomes slimmer and more elongated. 
On the right, we visualize the space of learned shapes by running PCA on all shapes obtained on training and test dataset. The three PCA axis visualized show interesting deformations. For example, in the motorcycle, the three visualized axes vary in prominence of front tire, size of fuel-tank and concavity of seat respectively. 

\subsubsection{Camera-multiplex visualization over time.}
\begin{figure}[h]
  \centering
%   \vspace{1em}
  \includegraphics[width=\textwidth]{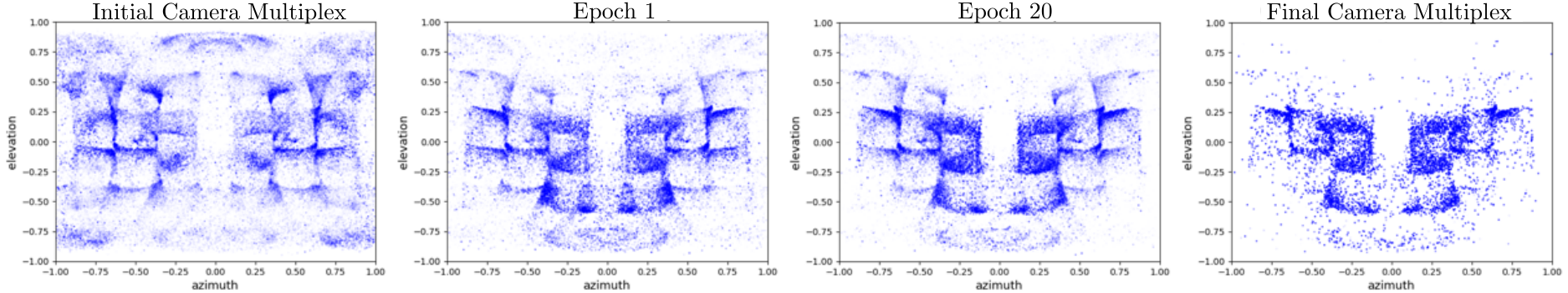}
  \caption{\small{{\bf Change in Camera Pose Distributions during training.}
  We show azimuth-elevation scatter plots of $K=40$ camera poses in all the camera-multiplex of the CUB train set as training progresses. Points corresponding to less probable cameras have a lower alpha value and are more transparent. Starting from top left, we have the camera poses after the camera-multiplex initialization, after 1 training epoch, after 20 training epochs and the final optimized camera poses. The number of camera poses in each multiplex ($K$) is pruned down from $40$ to $4$ after epoch 20.}}
%   \vspace{-1em}
  \label{fig:camera-dist-time}
\end{figure}

Figure~\ref{fig:camera-dist-time} shows how the azimuth-elevation distribution of camera poses in the camera-multiplex (over the entire training dataset) changes as training progresses. Observe that the distribution changes rapidly initially and results in a final distribution that is very different from the initial distribution.

\subsubsection{Qualitative comparison to CMR.}
We qualitatively compare results from U-CMR to 2 variants of CMR \cite{cmrKanazawa18}. The first is the official CMR model (CMR-official) that was trained using additional keypoint losses and used NMR \cite{NMR} as it's differentiable renderer. The second is our implementation of CMR (CMR-ours) which is similar to U-CMR in it's architecture for shape/texture, using Softras \cite{liu2019soft} for rendering, regularizing shape using the graph-laplacian and having the same template mesh as it's initial mean shape, but different from U-CMR in that it uses the ground-truth camera pose from SFM during training.  Unlike CMR-official, CMR-ours does not include vertex-keypoint reprojection loss. 

\begin{figure}[!h]
  \centering
  \includegraphics[width=\textwidth]{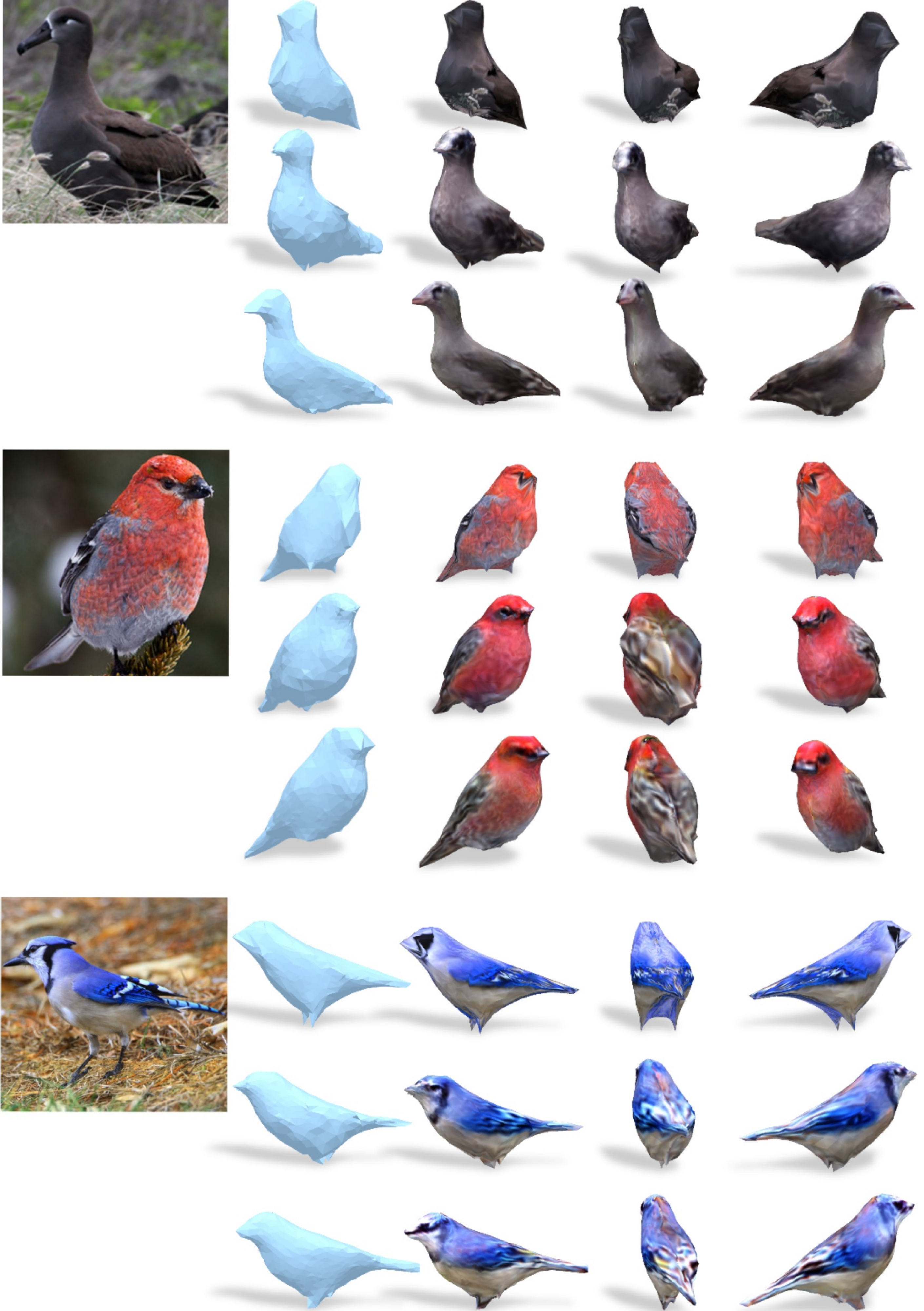}
  \caption{{{\bf CMR-official vs CMR-ours vs U-CMR.} We compare U-CMR (third row) to CMR-ours (second row) and CMR-official (first row) on selected images from the test dataset. The first 2 columns show the predicted shape and texture from the predicted camera viewpoint. The last 2 columns are novel viewpoints of the textured mesh.}}
  %\vspace{-1em}
  \label{fig:bird-cherry-1}
\end{figure}

\begin{figure}[!h]
  \centering
  \includegraphics[width=\textwidth]{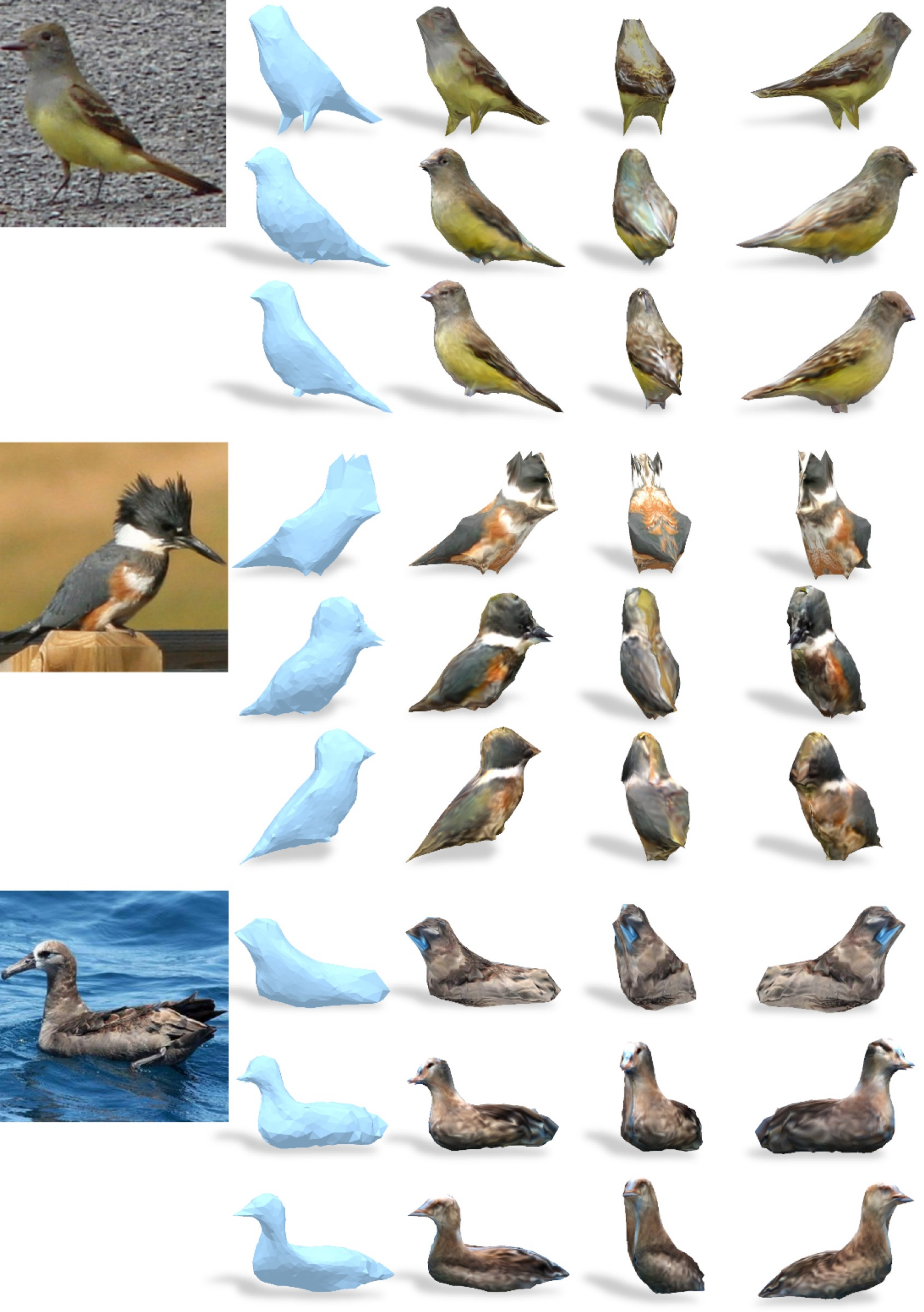}
  \caption{{{\bf CMR-official vs CMR-ours vs U-CMR.} We compare U-CMR (third row) to CMR-ours (second row) and CMR-official (first row) on selected images from the test dataset. The first 2 columns show the predicted shape and texture from the predicted camera viewpoint. The last 2 columns are novel viewpoints of the textured mesh.}}
  %\vspace{-1em}
  \label{fig:bird-cherry-2}
\end{figure}

In Figures~\ref{fig:bird-cherry-1}-\ref{fig:bird-cherry-2}, we compare CMR-official, CMR-ours and U-CMR. For each input image, the first row is from CMR-official, second is from CMR-ours and the last row is U-CMR. Observe that CMR-official is not as accurate as CMR-ours in capturing the shape and texture of the underlying bird but has pointier beaks and feet because of the keypoint reprojection loss it uses. The figures show that U-CMR shapes are qualitatively very similar to CMR-ours, hence exemplifying our assertion that U-CMR's camera-multiplex optimization alleviates the need for ground-truth camera pose supervision for most cases. %AK: Good paragraph! SG: Thank you! :)

\begin{figure}[!h]
  \centering
  \includegraphics[width=\textwidth]{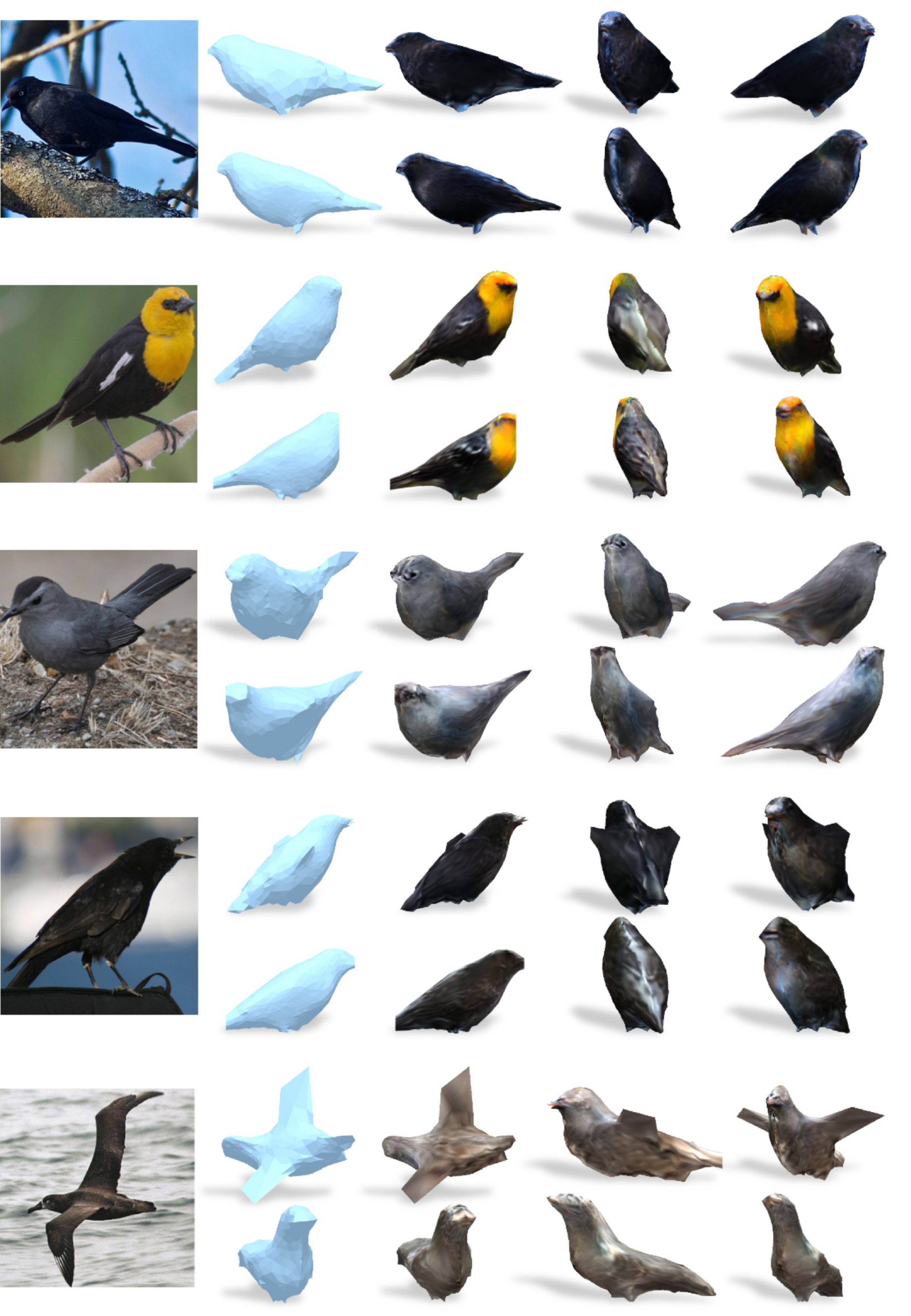}
  \caption{{{\bf CMR-ours vs U-CMR on \textit{random} subset.} We compare U-CMR (second row) to CMR-ours (first row) on a random subset of images from the testset. The first 2 columns show the predicted shape and texture from the predicted camera viewpoint. The last 2 columns are novel viewpoints of the textured mesh.}}
  %\vspace{-1em}
  \label{fig:bird-random-1}
\end{figure}

\begin{figure}[!h]
  \centering
  \includegraphics[width=\textwidth]{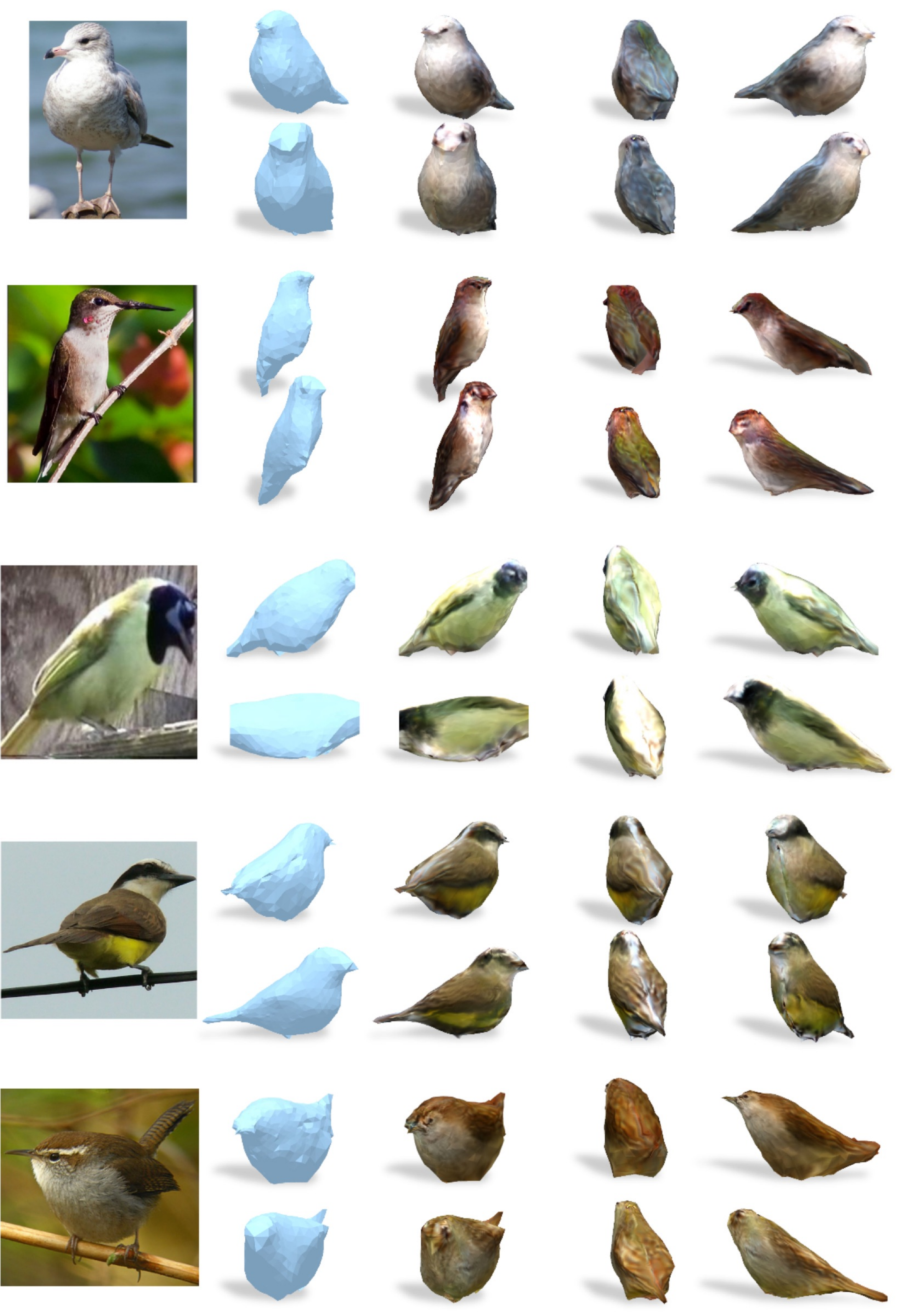}
  \caption{{{\bf CMR-ours vs U-CMR on \textit{random} subset.} We compare U-CMR (second row) to CMR-ours (first row) on a random subset of images from the test dataset. The first 2 columns show the predicted shape and texture from the predicted camera viewpoint. The last 2 columns are novel viewpoints of the textured mesh.}}
  %\vspace{-1em}
  \label{fig:bird-random-2}
\end{figure}

\begin{figure}[!h]
  \centering
  \includegraphics[width=\textwidth]{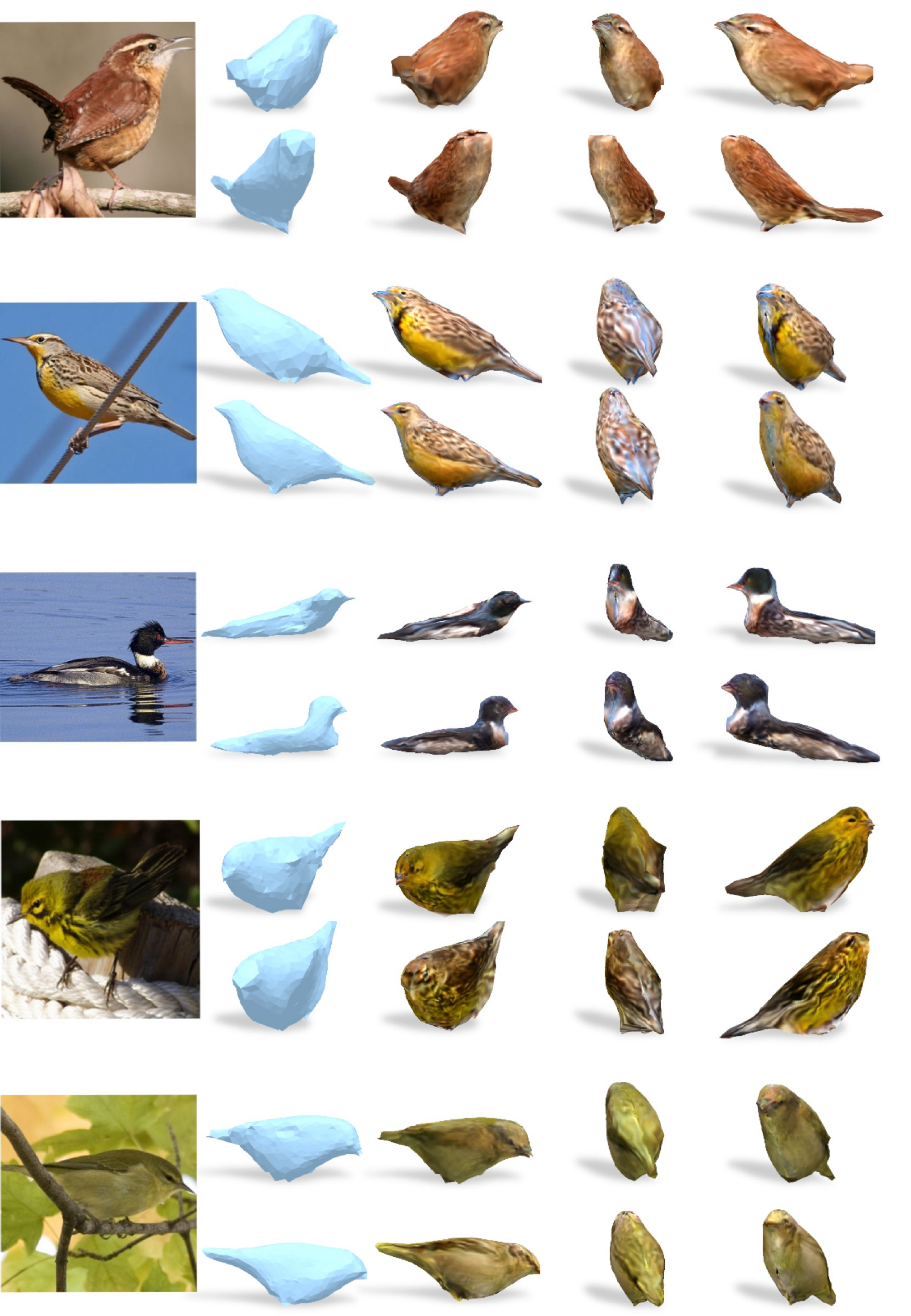}
  \caption{{{\bf CMR-ours vs U-CMR on \textit{random} subset.} We compare U-CMR (second row) to CMR-ours (first row) on a random subset of images from the test dataset. The first 2 columns show the predicted shape and texture from the predicted camera viewpoint. The last 2 columns are novel viewpoints of the textured mesh.}}
  %\vspace{-1em}
  \label{fig:bird-random-3}
\end{figure}

We compare U-CMR and CMR-ours on a random subset of $15$ images from the test dataset in Figures~\ref{fig:bird-random-1}-\ref{fig:bird-random-3}. Observe that U-CMR (second row) accurately predicts shapes that are very similar to those from CMR-ours when the bird is not articulating too much.

\subsubsection{Ablation on texture prediction model.}
\begin{figure}[h!]
  \centering
%   \vspace{1em}
  \includegraphics[width=\textwidth]{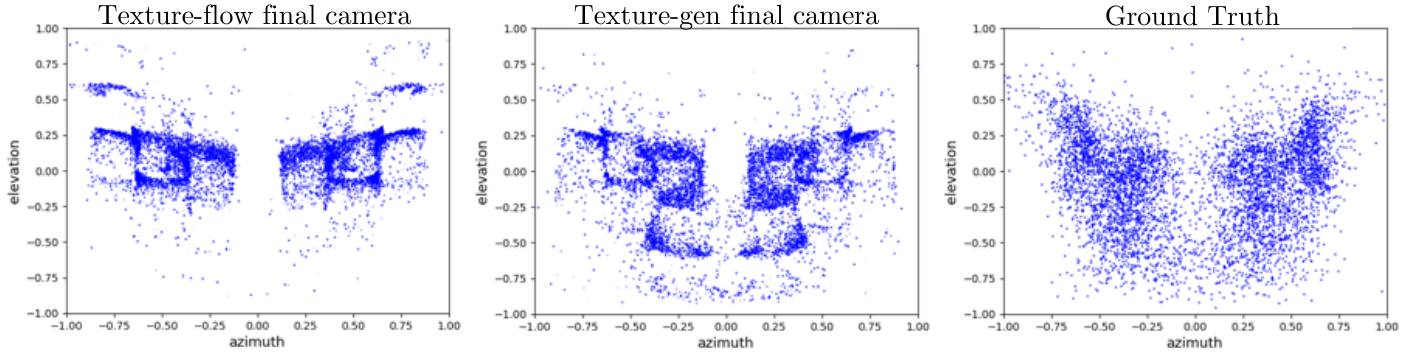}
  \caption{{{\bf Texture-flow camera-multiplex distribution on training set.} On the left, we show the azimuth-elevation distribution of the final camera-multiplex when we predict texture as a flow-field for sampling from the input image. Note how the camera poses have collapsed to 2 broad areas. Center: U-CMR camera-multiplex, Right: GT camera distribution
  }}
%   \vspace{-1em}
  \label{fig:texflow}
\end{figure}

% We promise this in main paper
We experiment with two different architectures for predicting the texture. First, we explore predicting texture as texture-flow, which is used in CMR. Texture-flow is a 2D positional offset for every pixel in the UV image that specifies where to sample the RGB values from the input image. %Most notable amongst was Texture-Flow wherein like CMR, we predict a flow-field into the input image and then sample it to get the RGB texture map.
Second, we predict the UV image values directly (Texture-gen) using a decoder attached to a bottleneck with spatial dimensions preserved. This is the final approach used in U-CMR. 
We observed that predicting a flow-field can lead to flat degenerate shapes and a collapse of optimized camera poses. Figure~\ref{fig:texflow} shows the collapse in the final optimized camera poses when predicting texture as a flow-field. This is because even when the camera pose is wrong, texture-flow is able to learn to adjust to the bad camera, as the output of the texture-flow across different instances is not necessarily correlated. However, when predicting the texture directly through a decoder, the network learns an implicit spatial prior of the texture across the dataset. For example, the network needs to learn to generate the texture of the eye at the same location in every texture map. Similarly for wings, breast, head etc\., as the texture map is predicted in a canonical semantic space. This spatial prior that is learned through a spatial decoder allows the texture prediction to disambiguate incorrect and correct poses in the camera-multiplex. 
% \sg{TODO if time permits: qualitative viz}

% We believe this is because while predicting a flow-field, the output of the network across different instances isn't necessarily correlated. However, when we predict texture as RGB, there are spatial priors that the network can learn. For example, the network needs to learn to generate the texture of the eye at the same location in every texture map. Similarly for wings, breast, head etc. as the texture map is predicted in a canonical semantic space.

% \subsubsection{Ablation on template shape}
% \sg{TODO}

% \subsubsection{Ablation on size of camera multiplex}
% \sg{TODO}

% \begin{figure}[h]
%   \centering
%   \includegraphics[width=\textwidth]{suppl_figures/FIG_cherry bird 3.pdf}
%   \caption{{{\bf CMR-official vs CMR-ours vs U-CMR on select images..} We compare U-CMR (third row) to CMR-ours (second row) and CMR-official (first row) on selected images from the test dataset. The first 2 columns show the predicted shape and texture from the predicted camera viewpoint. The last 2 columns are novel viewpoints of the textured mesh.}}
%   %\vspace{-1em}
%   \label{fig:bird-cherry-3}
% \end{figure}

\section{Training details}
%\subsection{Training details}
\subsection{Architecture details.}
Our code is available on our project page: \footnotesize{\url{https://shubham-goel.github.io/ucmr}}. The shape and texture predictor $f$ has an encoder-decoder architecture. Image $I \in \mathbb{R}^{256 \times 256 \times 3}$ is first encoded to latent feature map $z \in \mathbb{R}^{4 \times 4 \times 256}$ using Resnet-18. The shape head takes flattened $z \in \mathbb{R}^{16 \times 256}$ as input and passes it through 2 fully connected layers, each with $200$ output channels and then a final linear layer for predicting $\Delta_V \in \mathbb{R}^{|V| \times 3}$. % The texture head which takes as input a bilinearly interpolated $z \in \mathbb{R}^{4 \times 8 \times 256}$, comprises of 7 Resnet blocks with 256, 256, 256, 128, 64, 32, 16 output channels respectively. The feature maps undergo a $2\times$ bilinear upsampling after blocks 1, 3, 4, 5, 6. A final Conv layer operates on a $\mathbb{R}^{128 \times 256 \times 16}$ feature map and outputs texture map $I^{uv} \in \mathbb{R}^{128 \times 256 \times 3}$. 
The texture head takes the latent feature map and bilinearly samples it to $z \in \mathbb{R}^{4 \times 8 \times 256}$, this is followed by 7 Resnet blocks with 256, 256, 256, 128, 64, 32, 16 output channels respectively, with intermediate bilinear upsampling by a factor of 2 after blocks 1, 3, 4, 5, 6. This is then sent to a final convolution layer that outputs the texture map $I^{uv} \in \mathbb{R}^{128 \times 256 \times 3}$. The Resnet encoder uses ReLU activations while the shape and texture heads use Leaky-ReLU activations. All networks use batchnorm for normalization. We will release our code upon publication. 

After training the shape and texture prediction with camera-multiplex, we learn the feed-forward camera pose predictor $g$. We attach this as another head to the latent variable $z$ from the shared Resnet-18 trained for shape and texture prediction with camera-multiplex. We freeze the encoder, and then train a fully connected head for predicting camera scale $s \in \mathbb{R}$, translation $t \in \mathbb{R}^2$ and rotation (as quaternion $q \in \mathbb{R}^4$) through 2 fully connected layer each with 200 channels.
%comprises of the fixed pretrained encoder from $f$ and a trainable fully connected head for predicting camera scale $s \in \mathbb{R}$, translation $t \in \mathbb{R}^2$ and rotation (as quaternion $q \in \mathbb{R}^4$). 
% $g$ takes flattened $z \in \mathbb{R}^{16 \times 256}$ as input and passes it through 2 fully connected layers each with $200$ output channels and then a final linear layer for predicting the camera scale, translation and rotation quaternion.

% \subsubsection{More Implementation details.}
% \sg{todo} 

\end{document}